\documentclass[preprint,12pt,nopreprintline]{elsarticle}

\usepackage{amssymb}
\usepackage{amsmath}
\usepackage{booktabs}
\usepackage{multirow}
\usepackage{subcaption}
\usepackage{placeins}
\usepackage{hyperref}
\usepackage{cleveref}

\newcommand{\Alpha}{\mathrm{A}}
\newcommand{\diff}{\mathrm{d}}
\newcommand*{\proposed}{\texttt{RFlash}\@\xspace}
\renewcommand{\vec}[1]{\boldsymbol{#1}}
\usepackage{xspace}

\journal{Medical Image Analysis}

\begin{document}
\begin{frontmatter}

\title{Shadow Reduction in Ultrasound Imaging Using Differentiable Simulation and Radiance Field Decomposition}

\author[label1]{Valentin Bacher}
\author[label3,label1]{Pak Hei Yeung}
\author[label4,label5]{Bernhard Kainz}
\author[label1,label6]{Madeleine K Wyburd}
\author[label1]{Nicola K Dinsdale}
\author[label2]{Michael Gray}
\author[label1]{Ana IL Namburete}
\affiliation[label1]{organization={Oxford Machine Learning in NeuroImaging Lab}, addressline={Department of Computer Science}, city={University of Oxford}, postcode={OX1 3QD}, country={United Kingdom}}
\affiliation[label2]{organization={Institute of Biomedical Engineering, University of Oxford}, addressline={Marcela Botnar Wing}, city={Oxford}, postcode={OX3 7LD}, country={United Kingdom}}
\affiliation[label3]{organization={Quantitative Healthcare Analysis (qurAI) Group, Informatics Institute}, addressline={University of Amsterdam}, city={Amsterdam}, postcode={1098 XH}, country={The Netherlands}}
\affiliation[label4]{organization={Friedrich-Alexander-Universität Erlangen-Nürnberg}, addressline={Werner-von-Siemens Str. 61}, city={Erlangen}, postcode={91052}, country={Germany}}
\affiliation[label5]{organization={Imperial College London}, addressline={180 Queen’s Gate}, city={London}, postcode={SW7 2AZ}, country={United Kingdom}}
\affiliation[label6]{organization={Department of Computer Science, University of Copenhagen}, addressline={Universitetsparken 1}, city={Copenhagen}, postcode={2100}, country={Denmark}}

\begin{abstract}
Acoustic shadows from bone and other highly attenuating tissues obscure clinically important structures in ultrasound. 
In fetal brain imaging, skull-induced artefacts disproportionately degrade the hemisphere closer to the transducer (proximal), limiting symmetric assessment of the two hemispheres. 
Existing correction methods require raw scanner data, impose restrictive assumptions on tissue properties, or rely on generative models that may hallucinate anatomy. 
We present \proposed, a physics-informed post-processing method that decomposes beamformed ultrasound images into explicit attenuation and scatter-intensity maps using a differentiable radiance-field formulation of image formation. 
Attenuation-adaptive re-rendering then removes the dependence of the signal at each depth on the intervening tissue, equivalent to virtually advancing the transducer into the tissue. 
Across 1,261 3D fetal brain volumes, 143 real 2D curvilinear abdominal scans, and 1,200 simulated 2D linear-probe liver scans, \proposed reduces shadow-related intensity differences more effectively than classical Hughes–Duck attenuation correction

For a gestational-age model trained on the distal hemisphere (further from the transducer) and applied to the proximal hemisphere, prediction error decreases by 5.1 days (40\%) relative to the original images. 
The estimated attenuation maps also yield shadow-confidence maps that improve random-forest bone-shadow segmentation over the image alone and receive greater SHAP importance than an existing neural confidence-map baseline, suggesting greater physical consistency. 
\proposed requires neither hardware modification nor access to raw scanner data and supports 2D and 3D acquisitions with linear and curvilinear probes, making it widely applicable allowing clinicians to use our method on their already acquired scanners and images.

\end{abstract}

\begin{keyword}
Ultrasound imaging \sep Shadow reduction \sep Differentiable ultrasound simulation \sep Radiance fields \sep Fetal brain imaging \sep Explainable machine learning
\end{keyword}
\end{frontmatter}

\section{Introduction}
\label{sec:introduction}

Ultrasound is a standard imaging modality in prenatal care \citep{pogledic24prenatal}.
It is cost-effective, portable, and low risk, which makes it suitable for repeated clinical examinations.
Despite substantial improvements in image quality in recent decades, ultrasound images still suffer from low signal-to-noise ratio and physics-based artefacts, including acoustic shadows, reverberations, drop-out, and enhancement artefacts.
These artefacts reduce local contrast and can obscure structures that are clinically relevant \citep{nelson003dultrasound}.

\begin{figure}[!htbp]
\centering
    \includegraphics[width=\linewidth]{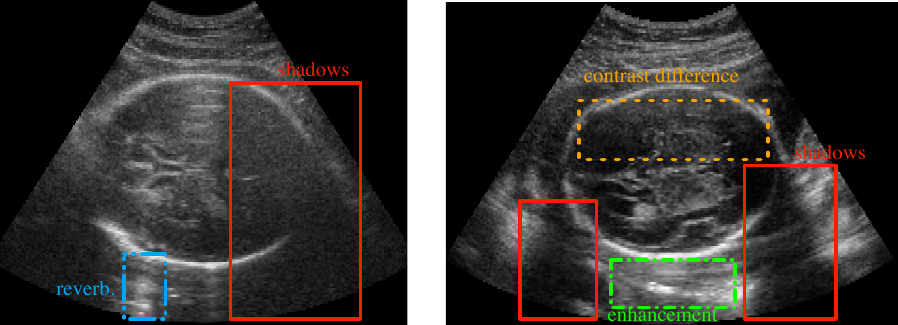}
\caption{Acoustic artefacts in clinical B-mode ultrasound. Solid red boxes mark acoustic shadows behind strongly attenuating structures. The dashed blue box marks reverberation, the dashed green box marks enhancement artefacts, and the dashed orange box marks the contrast difference between the proximal (closer to the transducer) and distal (further away from the transducer) fetal brain hemispheres. \proposed targets enhancement and shadow artefacts and, to some extent, the contrast difference between the hemispheres.}
\label{fig:introduction:artefacts}
\end{figure}

Shadow artefacts are particularly limiting when highly attenuating structures cannot be avoided by changing the probe position.
This is common in fetal brain ultrasound, where the skull increasingly calcifies during gestation and produces stronger attenuation.
As a result, the distal (further from the transducer) hemisphere is often better visible than the proximal (closer to the transducer) hemisphere, which limits assessment of bilateral anatomy and symmetry \citep{malinger20ultrasound,paladini21ultrasound}.
This is clinically relevant as brain symmetry is an important biomarker in fetal neurodevelopmental assessment \citep{namburete23normative}.
Representative examples of these artefacts are shown in \Cref{fig:introduction:artefacts}.
Effective shadow reduction could, therefore, increase the amount of usable image information by improving visibility in the proximal hemisphere.

Although shadow reduction is important, it remains an unsolved problem \citep{nelson003dultrasound,xu22shadowconsistent,hacihaliloglu17enhancement}.
Most ultrasound reconstruction pipelines assume similar sound-energy reduction across neighbouring scanlines. Heterogeneous tissue violates this assumption, which can lead to local under- or over-exposure \citep{aldrich07basic}.
Classical shadow-reduction methods compensate for such intensity variation using restrictive statistical or physical assumptions.
For example, \citet{hughes97automatic} and \citet{knipp97attenuation} assume a linear relation between attenuation and scattering, which may break down in complex geometries and heterogeneous tissue.
Frequency-based attenuation estimation has also been explored \citep{treece05ultrasound}, but requires frequency information in their RF data, which almost all commercially available scanners don't provide, limiting its use as a post-processing method.
Learning-based approaches can reduce artefacts visually, but may lack physical grounding, may generalize poorly across probes, and can introduce hallucinated image content \citep{yasutomi19shadow,yasutomi21shadow}.

\begin{figure}
    \centering
    \includegraphics[width=\linewidth]{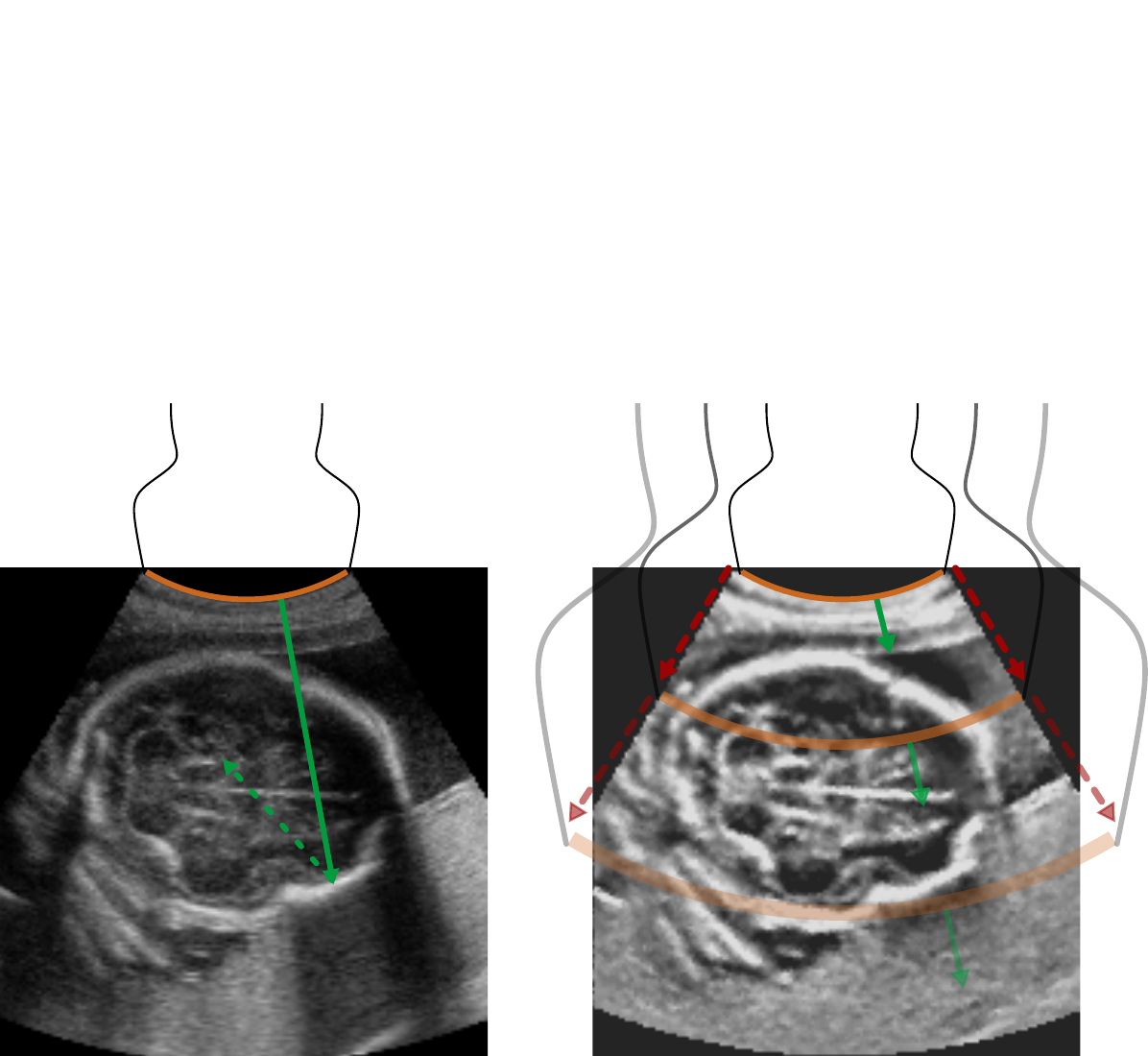}
    \caption{Principle of shadow-free rendering. Left: in a conventional acquisition, the intensity at a given depth (green ray) depends on all tissue between the transducer and that depth, so attenuating structures such as the skull cast shadows into everything below them. Right: \proposed renders each depth as though the transducer had been advanced to just above it (orange arcs, short green rays), making the rendered intensity independent of overlying tissue and suppressing the shadow.}
    \label{fig:introduction:teaser}
\end{figure}

In this work, we introduce \textbf{R}adiance \textbf{F}ie\textbf{l}d for \textbf{a}coustic \textbf{sh}adow reduction (\proposed), a physics-informed post-processing approach that explicitly models ultrasound propagation through tissue.
\proposed uses a fully differentiable ultrasound simulation pipeline to decompose ultrasound images into physically interpretable parameter maps.
These parameters are then used to render attenuation-corrected, shadow-reduced images.
Conceptually, the shadow-free rendering can be interpreted as virtually moving the transducer through the tissue, as illustrated in \Cref{fig:introduction:teaser}.
Each rendered depth is made less dependent on tissue closer to the original transducer position, preventing proximal structures from casting shadows onto deeper regions.
As the method operates as post-processing and uses a canonical scanline-based imaging model, it can be applied without hardware constraints and can support different acquisition geometries, including linear and curvilinear probes as well as 2D and 3D ultrasound.

We evaluate \proposed on fetal brain, real abdominal, and simulated liver ultrasound data spanning 3D acquisitions with a 3D probe, curvilinear 2D acquisitions, and simulated linear-probe data.
The method improves image quality both visually and quantitatively, producing more uniform contrast across fetal brain hemispheres and reducing shadow-related intensity differences in real images.
We further show that the recovered proximal-hemisphere information is relevant for a downstream fetal age-prediction task.
Finally, we derive shadow confidence maps directly from the attenuation parameter and show that these maps are informative for random-forest shadow segmentation and perform well compared to existing methods.

The code, as well as links to most of the data, can be found in the GitHub repository: \href{https://github.com/vbacher/RFlash}{https://github.com/vbacher/RFlash}

\section{Related Work}
\label{sec:literature-review}

\subsection{Shadow Artefact Reduction and Detection}

Earlier shadow reduction methods typically used single-step estimates, due to computational limitations. 
These approaches often estimate attenuation statistically using maximum-likelihood estimators \citep{melton83rational}, or derive closed-form attenuation and scatter maps. 
\citet{knipp97attenuation} and \citet{hughes97automatic} assume a linear relation between attenuation and scattering, which allows closed-form solutions but over simplifies tissue properties. 
Related ideas have also been used more recently in optical coherence tomography, which has similar echo-based imaging principles but uses coherent light \citep{girard11shadow, vermeer14depthResolved}. 
To the best of our knowledge, these are the only shadow-reduction techniques that work as a post-processing step on B-mode images and does not require any data annotation for training and/or inference, which is why we chose Hughes shadow reduction as our baseline. 
Frequency-based methods estimate attenuation from attenuation-dependent shifts in the signal centre frequency \citep{treece05ultrasound, kim08hybrid} and require frequency information only contained in Radio Frequency (RF) data, that most clinical systems do not provide.

More recent methods use iterative optimisation. \citet{yu10backscatter} solve a small partial differential equation in each iteration, which increases computational cost. 
Machine-learning approaches often focus on shadow detection rather than shadow removal, commonly using generative adversarial networks (GANs) \citep{yasutomi19shadow,yasutomi21shadow}. 
These methods can produce visually plausible results, but GANs may hallucinate image content, which limits their reliability for medical use, as hallucinations may miss abnormalities affecting diagnosis. 

\proposed operates on B-mode images and does not require frequency information or complex solvers at each iteration. 
It is trained on an individual-image basis, preventing knowledge transfer between subjects and reducing the risk of hallucinations.

\subsection{Radiance Fields}

Neural radiance fields (NeRFs) learn volumetric scene representations by mapping spatial coordinates and viewing directions to colour and opacity \citep{mildenhall21nerf}. This implicit representation has been widely extended across domains \citep{gao23nerf}. In medical imaging, radiance fields have been used to reconstruct 3D ultrasound volumes from 2D slices \citep{yeung24sensorless, wysocki24ultranerf, dagli24nerfus,gaits24ultrasound,hu24neural, zhang25hfusnerf}. Hybrid and Gaussian-splatting variants further reduce training time \citep{eid24rapidvol,guo24ulrenerf,eid25ultragauss}.

We build on this work by adapting Radiance Fields to ultrasound physics. Instead of learning visual appearance, our model learns tissue parameters with a differentiable, physics-informed renderer. Unlike conventional Radiance Fields trained from 2D projections, we use \emph{cross-sectional slice data} and treat each ray as a direct observation of local acoustic interactions. For efficiency and interpretability, we use a \emph{fully explicit representation} \citep{chen22tensorf, chan22efficient}, which supports fast training and transparent parameter estimates.

To our knowledge, this is the first use of radiance fields for shadow artefact reduction in ultrasound.

\subsection{Ultrasound Simulation}

Our scene representation is learned from an ultrasound slice or volume using a differentiable ultrasound simulation. Efficient ultrasound simulators use convolutions or point-spread functions with tissue parameters or scatterers \citep{bamber80ultrasonic,meunier95ultrasonic,burger08simulation,hergum09fast,marion09toward}, and can be implemented efficiently on GPUs \citep{mattausch18imagebased,gjerald12realtime}. These methods approximate interference patterns from point scatterers well, but do not explicitly model shadow formation along scanlines. Ray-tracing and scanline-tracing methods address this by modelling acoustic propagation at tissue interfaces \citep{wein08automatic,law11ultrasound}.

Extensions with convolutional scattering models \citep{burger13realtime, salehi15patient} or Monte Carlo sampling \citep{mattausch16monteCarlo, mattausch18realistic} improve realism, but are computationally expensive and often non-differentiable. This limits their use in learning-based optimisation. Recent differentiable ray-based models improve optimisation compatibility \citep{duelmer2025ultraray}, but still rely on restrictive sampling.

In this work, we use a simplified differentiable ray-based model that ignores cross-scattering effects \citep{bertramo25diffus} and models signal intensities directly, which is in line with other physics models used for shadow reduction and 2D to 3D reconstruction \citep{hughes97automatic, wysocki24ultranerf}.
This avoids point-spread-function convolution and enables fast, physically consistent simulation for iterative optimisation and shadow correction.

\section{Methods}
\label{sec:methods}

\begin{figure}[ht]
\centering
\includegraphics[width=\linewidth]{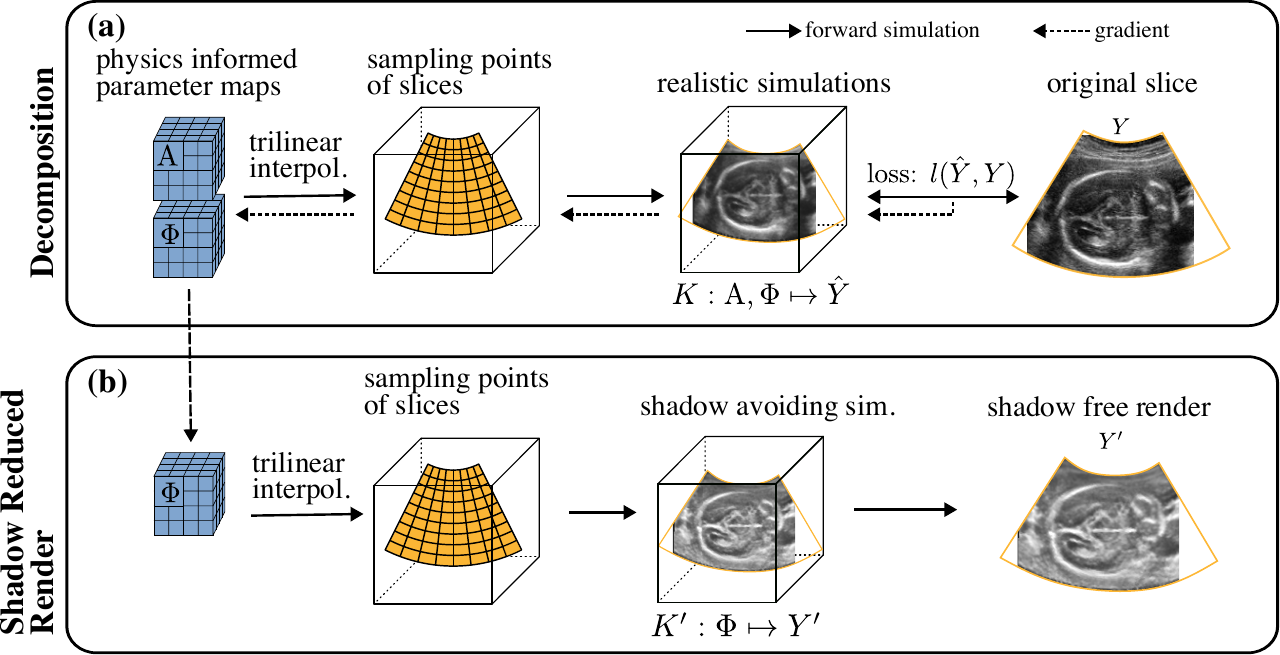}
\caption{The two phases of \proposed. (a) Decomposition: attenuation $\Alpha$ and scatter intensity $\Phi$ are stored in an explicit volume, sampled along scanlines by trilinear interpolation, and passed through the differentiable forward simulator $K: \Alpha,\Phi \mapsto \mathrm{\hat{Y}}$. The rendered slice $\mathrm{\hat{Y}}$ is compared with the original slice $\mathrm{Y}$ by the loss $l(\mathrm{Y},\mathrm{\hat{Y}})$, and gradients (dashed arrows) update the parameter volumes. (b) Shadow-reduced rendering (bottom): the converged $\Phi$ from (a) is re-rendered through the shadow-avoiding operator $K': \Phi \mapsto \mathrm{Y'}$, which omits the attenuation term and so produces a shadow-free image. Solid arrows: forward simulation. Dashed arrows: gradient flow.}
\label{fig:methods:model}
\end{figure}

Analogous to most NeRF-like models \citep{mildenhall21nerf}, our method comprises two main components: a scene representation learned as a set of physics-informed tissue parameters and a differentiable ultrasound simulation that renders those parameters into ultrasound images. Both components are used in two steps: decomposition (a), which learns the tissue parameters, and shadow-reduced rendering (b), which renders the shadow-free images, as shown in \Cref{fig:methods:model}.

\subsection{Ultrasound image formation model}

\begin{figure}[ht]
\centering
    \includegraphics[width=\linewidth]{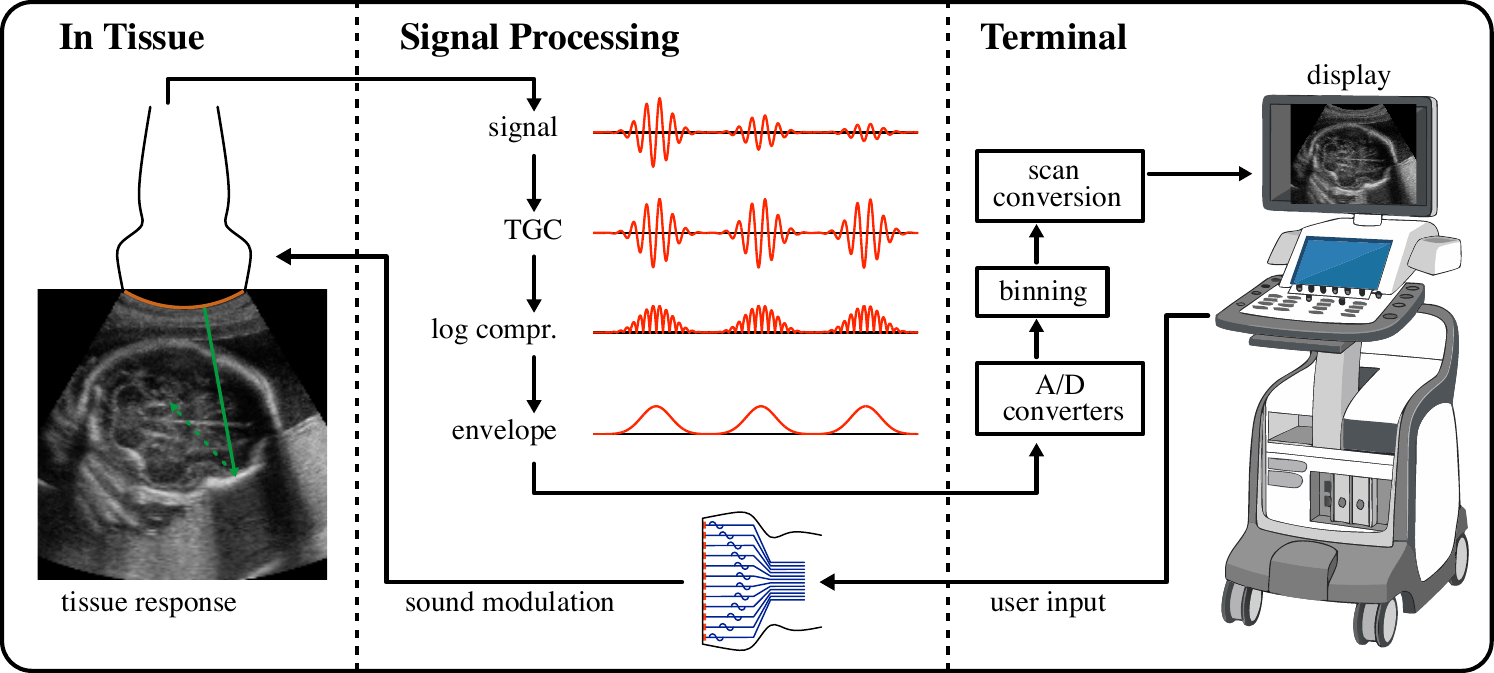}
\caption{The canonical ultrasound imaging pipeline on which \proposed is built. Tissue response (left) is modulated by the propagating sound field, then processed in the scanner (centre: envelope detection, time-gain compensation, log compression) and converted to a display image (right: A/D conversion, binning, scan conversion). \proposed operates on the displayed, beamformed image and inverts the shaded operations. A/D conversion and binning are lossy and are not modelled. Compare with \Cref{fig:methods:simulation}, which shows the corresponding differentiable pipeline for beamformed images.}
\label{fig:methods:sig_flow}
\end{figure}

Ultrasound scanners differ in their proprietary acquisition and processing pipelines. 
Since \proposed is applied as a post-processing method to the final B-mode image, we assume that all images were generated using the standard pipeline shown in \Cref{fig:methods:sig_flow}. The exact signal processing pipeline of the image to be shadow removed does not need to be known for \proposed to work.

\begin{figure}
\centering
\includegraphics[width=.3\linewidth]{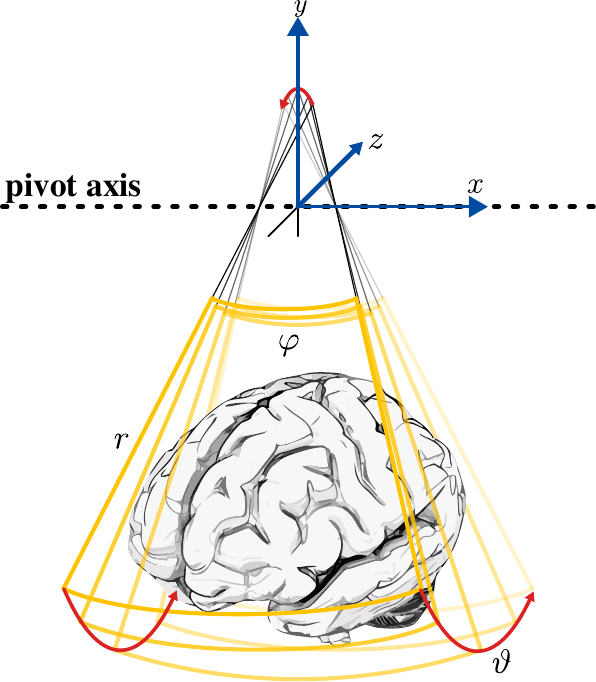}
\caption{Geometry of the 3D acquisition. A mechanically swept probe is modelled as a curvilinear 2D array (yellow fan) rotated about an internal pivot axis (dashed), so that a voxel is addressed by depth $r$ and two angular coordinates: $\varphi$ within the fan and $\vartheta$ about the pivot. Each scanline is treated as an independent propagation path.}
\label{fig:methods:geometry}
\end{figure}

We assume that the input data have already been beamformed at the transducer, which is the process of focusing the ultrasound beam and forming the initial image. This allows each scanline to be treated independently along the propagation direction. The measured signal is corrected by time-gain compensation (TGC), logarithmically compressed, and envelope-detected using a Hilbert transform before display. It is then scan-converted from samples along scanlines into a Cartesian image grid. For linear probes, scan conversion reduces to resampling anisotropic scanline samples onto image pixels. For curvilinear probes, polar coordinates are additionally mapped to a Cartesian grid. For 3D probes, a mapping from the scanline--depth coordinate system to Cartesian volumes for storage and display is needed, which requires some knowledge of the 3D probe type. The 3D data we used were acquired with a mechanical probe that we model as a curvilinear 2D probe rotated around an internal probe axis, producing different angular coordinates in the two lateral directions, as illustrated in \Cref{fig:methods:geometry}. We do not model analogue-to-digital (A/D) conversion or binning, because the available inputs are already digitised and these lossy steps cannot be recovered from post-processed images.

\subsection{Explicit radiance-field representation}

Our model follows the radiance-field view of differentiable image formation, as summarised in \Cref{fig:methods:model}. Unlike the original NeRF formulation \citep{mildenhall21nerf} and recent ultrasound radiance-field methods \citep{yeung24sensorless,wysocki24ultranerf}, we do not represent the scene with a multi-layer perceptron (MLP). MLPs provide a continuous representation, but are slow to optimise and less interpretable because each weight can affect the scene globally. In our setting, the input volumes have a fixed discrete resolution; therefore, an implicit continuous representation provides limited practical benefit.

We instead use a fully explicit volumetric representation. This representation is straightforward to initialise, inspect, and optimise, at the cost of a larger memory footprint. During training, each pixel in a reference ultrasound image is mapped into the representation space, which has the dimensions of the padded input image. The corresponding tissue parameters, such as attenuation and scatterer density, are sampled from the explicit volume using trilinear interpolation and passed through the differentiable ultrasound simulator. The rendered 2D image is then compared with the reference image to update the volumetric parameters.

\subsection{Differentiable ultrasound simulation}

The simulator uses two learned material parameters: attenuation $\Alpha$, which captures local effects that reduce image intensity, and scatter intensity $\Phi$, which controls the local backscattered signal. 
For numerical stability, attenuation is scaled by a constant factor $g$, which avoids very small gradients during optimisation. 
We do not separately estimate reflection and refraction, because most structures are observed from a single acoustic view and these effects cannot be uniquely disentangled from the available images. 
We also assume a constant speed of sound $c$, yielding a linear relation between imaging depth and time of flight, $t \propto \vec{x}$, which is standard for B-mode ultrasound \citep{postema11diagnostic}.

Following \citet{hughes97automatic}, we model the simulated signal before scan conversion as the product of scatter intensity and the remaining acoustic intensity after attenuation:
\begin{equation}
 \mathrm{\tilde Y} (\vec{x}) = \Phi (\vec{x}) \mathrm{I}(\vec{x}).
\label{equ:methods:base_model}
\end{equation}
The remaining intensity $\mathrm{I}(\vec{x})$ follows an exponential decay based on the attenuation coefficient $\Alpha$ along the propagation path, analogous to the Beer-Lambert law:
\begin{equation}
 \mathrm{I}(\vec{x}) = \mathrm{I}_0 \exp\left(-\int_0^{\vec{x}} \Alpha(\vec{u}) g\, \diff \vec{u}\right).
\label{equ:methods:attenuation}
\end{equation}
Substitution gives the measured pre-display signal
\begin{equation}
\mathrm{\tilde Y} (\vec{x}) = \mathrm{I}_0 \Phi (\vec{x}) \exp\left(-\int_0^{\vec{x}} \Alpha(\vec{u})g \, \diff \vec{u}\right).
\label{equ:methods:measured}
\end{equation}
Here, $\mathrm{\tilde Y}(\vec{x})$ denotes the envelope-detected backscattered intensity after acoustic propagation and attenuation, but before TGC and log compression. This formulation is consistent with classical analytic ultrasound models used for shadow reduction \citep{hughes97automatic,yu10backscatter}.

\begin{figure}
    \centering
    \includegraphics[width=\linewidth]{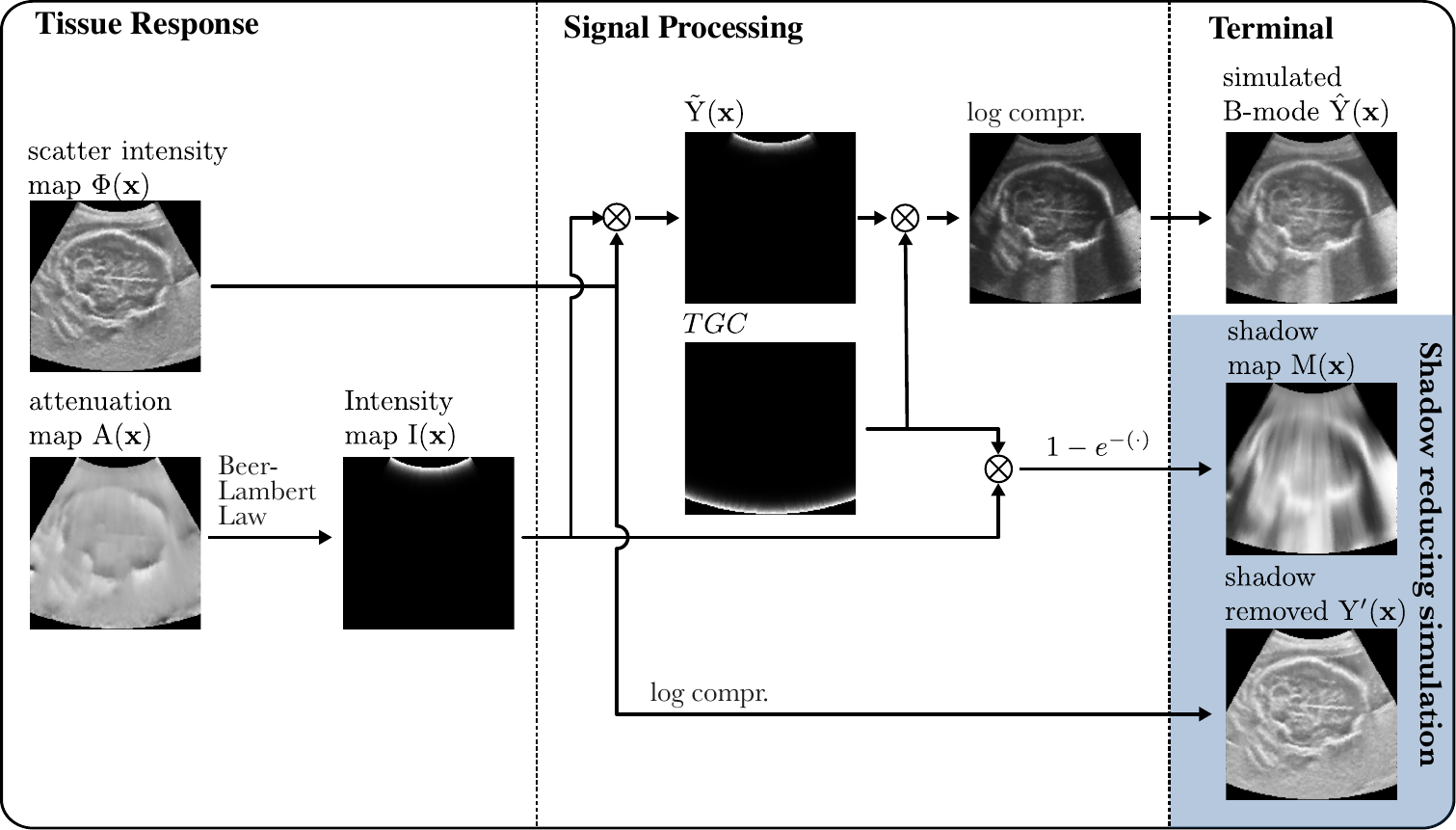}
    \caption{Visual representation of the differentiable simulation pipeline as defined by \cref{equ:methods:base_model}--\cref{equ:methods:shadow_mask_normalized}. The first target (top right) is the simulated ultrasound image $\mathrm{\hat{Y}}$, while the lower two targets are the shadow-reduced image $\mathrm{Y'}$ and the shadow map $\mathrm{M}$, which are only relevant after the tissue parameters have been estimated (b in \Cref{fig:methods:model}).}
    \label{fig:methods:simulation}
\end{figure}

Because the backscattered signal decays with depth, clinical ultrasound systems apply TGC before log compression to maintain approximately uniform contrast across the image. The scanner-specific TGC profile is typically not stored with the image data. We therefore model TGC as a depth-dependent exponential gain,
\begin{equation}
TGC (\vec{x}) = \exp (c(\vec{x})),
\label{equ:methods:tgc}
\end{equation}
where $c(\vec{x})$ is linear along the propagation direction. Equivalently, this can be written as
\begin{equation}
    c(\vec{x}) = \int_0^{\vec{x}} \Alpha \, \diff \vec{u} .
\label{equ:methods:tgc_exponent}
\end{equation}
After applying TGC and log compression, the simulated B-mode intensity is
\begin{equation}
 \mathrm{\hat Y} (\vec{x}) = \frac{\log (1+\lambda \left[TGC (\vec{x}) \mathrm{\tilde Y} (\vec{x})\right])}{\log (1+\lambda )},
\label{equ:methods:simulation}
\end{equation}
where $\lambda$ controls the dynamic range. Equations~\eqref{equ:methods:base_model}--\eqref{equ:methods:simulation} define the forward simulation operator $K:\Alpha, \Phi \mapsto \mathrm{\hat Y}$, shown in \Cref{fig:methods:model}. The corresponding differentiable processing pipeline is shown in \Cref{fig:methods:simulation}.

\subsection{Shadow-reduced rendering}

Shadow artefacts arise when a globally defined TGC profile does not compensate for spatially varying attenuation, especially behind strongly reflecting or absorbing structures such as bone. Shadow reduction can, therefore, be formulated as estimating a structure-aware compensation profile from the learned attenuation map. Instead of applying a global gain, we compute the adaptive gain
\begin{equation}
TGC_{sr} (\vec{x}) = \exp\left(\int_0^{\vec{x}} \Alpha(\vec{u}) g \, \diff \vec{u}\right).
\label{equ:methods:tgc_sr}
\end{equation}
Applying this compensation cancels the attenuation term in \Cref{equ:methods:measured} and yields the shadow-reduced rendering
\begin{equation}
 \mathrm{Y'} (\vec{x}) = \frac{\log (1+\lambda \left[\mathrm{I}_0 \Phi(\vec{x})\right])}{\log (1+\lambda )},
\label{equ:methods:sim_sr}
\end{equation}
which defines the shadow-reduced simulation operator $K': \Phi \mapsto \mathrm{Y'}$.

As shown in \Cref{fig:methods:model}, the attenuation and scatter maps are first optimised from the available training slices using the realistic forward operator $K$. Once these material maps have been estimated, the shadow-reduced operator $K'$ renders the final shadow-reduced image.

\subsection{Shadow confidence maps} 
\label{subsec:methods:shadow_mask}

We derive a shadow confidence map by comparing the measured image with its shadow-reduced rendering in the linear intensity domain. Let $L(\cdot)$ denote the log-compression operator in \Cref{equ:methods:simulation}. Assuming $\mathrm{I}_0=\mathrm{I}$, where $\mathrm{I}$ is the identity, the attenuation-related intensity ratio is
\begin{equation}
\begin{split}
\mathrm{\tilde M}(r)
&= TGC(r) \mathrm{I}(r) \\
&= \frac{TGC(r)\Phi(r) \mathrm{I}(r)}{\Phi(r)} \\
&= \frac{L^{-1}(\mathrm{Y}(r))}{L^{-1}(\mathrm{Y'}(r))}.
\end{split}
\label{equ:methods:shadow_mask}
\end{equation}

Here, $\mathrm{Y}(r)$ is the measured image and $\mathrm{Y'}(r)$ is the corresponding shadow-reduced rendering. The normalised shadow confidence map is then obtained as

\begin{equation}
\mathrm{M}(r) = 1 - \exp\left[-\mathrm{\tilde M}(r)\right].
\label{equ:methods:shadow_mask_normalized}
\end{equation}

\section{Experimental Setup}
\label{sec:experiments}
\subsection{Datasets}
\label{subsec:exp_setup:datasets}
For evaluation, we used four different data sources spanning three different transducer geometries.

The main dataset used is a set of 3D \textbf{fetal-head ultrasound volumes} acquired as part of the INTERGROWTH-21$^\textrm{st}$ study~\cite{villar14likeness}, one of the largest fetal head ultrasound studies. 
Data were collected using a Philips HD9 (Philips Healthcare, Amsterdam, The Netherlands) scanner equipped with a curvilinear abdominal transducer (V7-3 3D). 
The dataset comprised 1261 volumes from fetuses between 15 and 31 gestational weeks (GW), each standardised to a spatial resolution of $160 \times 160 \times160 $ pixels. 
Further details on dataset acquisition, selection criteria, and preprocessing are provided in \cite{namburete23normative}.

We also evaluated our method on a publicly available \textbf{abdominal ultrasound dataset} \citep{orlando2022ussimandsegm}. We used only the real scans from the test dataset, which contains 213 different 2D ultrasound scans of the liver, spleen, and kidneys. For 70 of these scans, no geometry could be determined by our geometry estimator, mostly because of coupling artefacts between the transducer and the skin. Without knowing a good estimation of the scanner geometry, we cannot accurately apply our shadow reduction method. As our method focuses on shadow reduction rather than geometry estimation, we omitted these scans.

To evaluate our method against the one used in \citet{wysocki24ultranerf}, we used their \textbf{simulated liver dataset} containing 1200 simulated liver ultrasound scans from a linear probe.

To evaluate the shadow maps, we used 14 2D curvilinear \textbf{fetal scans} and their shadow maps for which the scanner geometry could be determined from \citet{meng18automatic}.

Links to the publicly available data can be found in the GitHub repository: \href{https://github.com/vbacher/RFlash}{https://github.com/vbacher/RFlash}

\subsection{Baseline shadow reduction}

As a baseline, we used the shadow reduction proposed by \citet{hughes97automatic}, as its approach can be adapted for use in post-processing and is the basis of most classical shadow-reduction techniques. In the original paper, shadow reduction was applied directly to radio-frequency (RF) data and, therefore, requires minor adaptation for use as a post-processing method.

The problem formulation used in their study is the same as Equation \eqref{equ:methods:measured}. Using their assumption that scattering intensity and attenuation are linearly dependent, \emph{i.e.}, $\Alpha = a + b \Phi$, with $a$ and $b$ as scalar factors, they solve Equation \eqref{equ:methods:measured} for $\Phi$, resulting in 

\begin{equation}
    b \Phi_{\varphi, \vartheta}(r) = \frac{\mathrm{\tilde Y}_{\varphi, \vartheta}(r) \exp (ar)}{ \int_r^{\infty}\mathrm{\tilde Y}_{\varphi, \vartheta}(\nu) \exp (a\nu) \, \diff \nu }\, .
    \label{equ:expset:hough_orig}
\end{equation}

\noindent Since $a$ is fitted to the data, the inverse log compression
\begin{equation}
L^{-1}(\mathrm{Y}_{\varphi, \vartheta}(r)):=\frac{(\lambda + 1)^{\mathrm{Y}_{\varphi, \vartheta}(r)}-1}{\lambda}= TGC(r) \mathrm{\tilde Y}_{\varphi, \vartheta}(r),
\label{equ:expset:inv_logcomp}
\end{equation}

\noindent can be used to find $a$ such that $TGC(r) = \exp (ar)$. Using this equivalence, \eqref{equ:expset:hough_orig} can be reformulated as

\begin{equation}
    b \Phi_{\varphi, \vartheta}(r) = \frac{L^{-1}(\mathrm{Y}_{\varphi, \vartheta}(r))}{ \int_r^{\infty}L^{-1}(\mathrm{Y}_{\varphi, \vartheta}(\nu)) \diff \nu }\, ,
    \label{equ:expset:hughes_II}
\end{equation}

\noindent allowing us to directly use the sampled slice $\mathrm{Y}_{\varphi, \vartheta}(r)$ after undoing log compression.
The shadow-reduced volumes using \eqref{equ:expset:hughes_II} are the benchmark we refer to as Hughes shadow reduction.

\subsection{Implementation details}
\label{subsec:experiments:implementation}

We initialised the parameter maps with a mean value of $1.2$ for $\Alpha$ and $4.25$ for $\Phi$. Further details on this choice are provided in \Cref{apdx:B:sens:init}. For stable training, we normalised the input image to the same mean value as $\Phi$, that is, $\mu = 4.25$ and a standard deviation of $\sigma = \mu/3 \approx 1.4167$. The constant offset $g=80$ was chosen to match the initialization of $\Alpha$, which also defines the TGC as $\exp(\int_0^{\vec{x}} \Alpha_0 g\,\diff \vec{u})$ with $\Alpha_0 = 1.2$.
For training, we used a weighted SSIM \citep{wang04ssim} and $L_2$ loss, with $1-SSIM(\mathrm{Y}, \mathrm{\hat Y})$ weighted by $0.95$ and $\lVert \mathrm{Y} - \mathrm{\hat Y}\rVert_2^2$ weighted by $0.05$.
The model was implemented in PyTorch \citep{paszke19pytorch} and optimised using Adam \citep{kingma17adam} with a learning rate of $0.03$ for $300$ epochs. Learning the representation took less than one minute on an NVIDIA GeForce RTX 4090 GPU with 24~GB of memory for each 3D fetal brain volume.

\subsection{Evaluation: Measurement of shadows}\label{subsec:experiments:measure}

We define an image $Im$ as strongly affected by shadows if the mean grey value in shadowed regions differs from that in regions not affected by shadows. Let $\Omega$ denote the set of pixels in $Im$ and let $\mathcal{M}\subseteq\Omega$ denote the thresholded shadow map obtained from Hughes shadow reduction or \proposed, using the procedure described in \Cref{subsec:methods:shadow_mask}. We denote the erosion operator by $\mathcal{E}$. Applying it to the thresholded shadow map and its complement defines the shadowed and non-shadowed regions as
\begin{equation}
    \mathcal{S}=\mathcal{E}(\mathcal{M}),
    \label{equ:expset:shadow_measure_regions}
\end{equation}
and the corresponding non-shadowed region as
\begin{equation}
    \mathcal{N}=\mathcal{E}(\Omega\setminus\mathcal{M}).
    \label{equ:expset:shadow_measure_nonshadow}
\end{equation}
Once these regions are identified, we equalised the image histograms to make the absolute values comparable. The difference in mean grey values is then
\begin{equation}
    \Delta\mu(Im)=\mu_{\mathcal{N}}(Im)-\mu_{\mathcal{S}}(Im),
    \label{equ:expset:shadow_measure_difference}
\end{equation}
so that positive values indicate higher mean intensity in the non-shadowed region than in the shadowed region. In contrast to \citep{girard11shadow}, we calculate the mean over the entire image, not just single layers. For visual and numerical evaluation of the methods, we used all available datasets apart from the 14 samples provided by \citep{meng18automatic}, as their statistical power was too small.

\subsection{Evaluation: Age prediction as clinical downstream task}\label{subsec:experiments:clinical_significance}

To assess the clinical relevance of the recovered structures in the proximal hemisphere, we evaluated whether a standard age-prediction model can make better use of this region after shadow reduction. Following \citep{wyburd21assessment,hesse24prototype}, we trained a 3D ResNet-10 \citep{he16deep} to predict fetal age from the better-visible distal hemisphere. The model was trained on the fetal brain volumes described in \Cref{subsec:exp_setup:datasets}, aligned to a common space and size-normalised, which reduces the possibility that age is predicted from global size differences rather than anatomical structure. In total, 1261 samples were used with a subject-level train, validation, and test split of $[868, 133, 260]$. Testing was then performed on both hemispheres to determine whether shadow reduction improves generalisation from the distal to the proximal hemisphere.

\subsection{Evaluation: Consistency of shadow maps}\label{subsec:experiments:mask_consistency}

To evaluate the physical consistency of the predicted shadow maps, and to compare them with the more widely studied shadow confidence maps, we compared our method with Ultra-NeRF \citep{wysocki24ultranerf}, which addresses a related problem in 2D to 3D ultrasound reconstruction. We first compared the resulting shadow confidence maps visually on fetal brain examples from \citet{meng2019weakly} and on the synthetic liver dataset of \citet{wysocki24ultranerf}. We then compared how informative the maps are for a downstream task by using a random forest to segment bone shadows, with the corresponding confidence map provided as an additional input. For a direct comparison, we followed the evaluation setup and, therefore, also the model choice and datasets of \citet{yesilkaynak2024ultrasound}. Further design details can be found in that work. The random forest was trained either on the simulated ultrasound images alone or on the ultrasound images together with either the Ultra-NeRF confidence maps or our confidence maps. Performance was evaluated numerically and by analysing the feature importance assigned to the shadow maps.

\section{Results}
\label{sec:results}

Visual results for three representative examples from each dataset are shown in \Cref{fig:results:vis_results}, with regions of interest highlighted by orange arrows. This figure also demonstrates shadow reduction on 3D, curvilinear 2D, and linear ultrasound imaging geometries.
\proposed effectively reduced shadow artefacts across the evaluated datasets.
In the fetal brain examples, structures in the proximal hemisphere, such as the Sylvian fissure, became more visible where sufficient image information was available for reconstruction.
The contrast between the proximal and distal hemispheres was also more uniform after shadow reduction.
Compared with Hughes shadow reduction, \proposed handled enhancement artefacts more consistently and better equalised contrast across regions with different intensity levels, as highlighted in \Cref{fig:results:vis_results}.

\subsection{Numerical Results}
\label{subsec:results:numerical_results}

\begin{figure*}[!htbp]
    \centering
    \includegraphics[width=\linewidth]{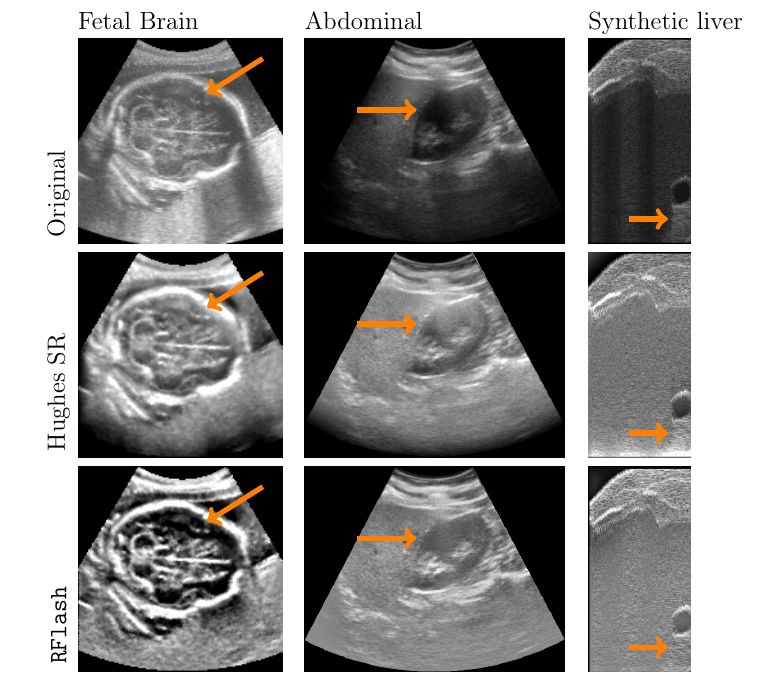}
    \caption{Qualitative comparison across acquisition geometries: fetal brain (3D mechanically swept), abdominal (curvilinear 2D) and simulated liver (linear 2D). Rows: original image, Hughes shadow reduction, \proposed. Orange arrows mark regions of interest: in the fetal brain, recovery of the Sylvian fissure in the proximal hemisphere. In the abdominal, over-brightening of shadowed regions by Hughes reduction, producing cloudy artefacts that \proposed avoids. In the liver dataset, enhancement artefacts, Hughes method does not remove.}
    \label{fig:results:vis_results}
\end{figure*}

The numerical comparison in \Cref{tab:exp:numres_dataset} shows that \proposed significantly outperformed Hughes shadow reduction on all real datasets.
For each method, the measurement was computed using the corresponding shadow map. Additional results for crossed shadow maps are provided in \Cref{apdx:A:additional_numerical_results}.
On the simulated dataset, Hughes shadow reduction achieved the best numerical score, possibly because the assumed linear relation between attenuation and scattering is better satisfied in simulated images than in real ultrasound data.
However, this metric does not capture enhancement artefacts, for which \proposed produced visually stronger results on the same dataset as marked in \Cref{fig:results:vis_results}.

\begin{table}[!htbp]
\caption{Mean intensity difference $\Delta\mu$ across datasets for the original ($O$), Hughes shadow-reduced ($H$), and \proposed ($SR$) volumes. Values are shown as mean $\pm$ standard deviation. The two right-most columns report the subject-wise differences relative to \proposed. Differences marked with $^*$ are significantly different from 0 in a t-test with a p-value $<\frac{0.01}{6}=0.001667$ for multiple-comparison correction.}
\label{tab:exp:numres_dataset}
\centering
\resizebox{\linewidth}{!}{%
\begin{tabular}{@{}l|ccc|cc@{}}
\toprule
Dataset & $\Delta\mu_O \downarrow$ & $\Delta\mu_H \downarrow$ & $\Delta\mu_{SR} \downarrow$ & $\left(\Delta\mu_O - \Delta\mu_{SR}\right)$ & $\left(\Delta\mu_H - \Delta\mu_{SR}\right)$ \\
\midrule
Syn liver & $64.12 \pm 27.21$ & $\mathbf{13.61 \pm 7.37}$ & $16.93 \pm 17.47$ & $47.19 \pm 20.06^*$ & $-3.32 \pm 18.40^*$ \\
Abdomen & $87.65 \pm 16.84$ & $13.41 \pm 10.13$ & $\mathbf{8.04 \pm 6.41}$ & $79.61 \pm 18.41^*$ & $5.37 \pm 10.33^*$ \\
Fetal Brain & $129.04 \pm 7.59$ & $48.26 \pm 22.84$ & $\mathbf{41.94 \pm 11.73}$ & $87.10 \pm 12.57^*$ & $6.32 \pm 29.55^*$ \\
\bottomrule
\end{tabular}
}
\end{table}

\subsection{Age Prediction as Clinical Downstream Task}
\label{subsec:results:age_pred}

As described above, age prediction was trained on size-normalised distal hemispheres and evaluated separately on the distal and proximal hemispheres.
The results are reported in \Cref{tab:exp:ageprednum}.
\proposed improved prediction accuracy for both hemispheres, including the distal hemisphere used during training.
The largest benefit was observed in the proximal hemisphere, which was more strongly affected by shadows.
Compared with the original images, \proposed reduced the prediction error in the proximal hemisphere by approximately 5 days relative to no shadow reduction and almost 2 days relative to Hughes shadow reduction, corresponding to relative improvements of 40\% and 19\%, respectively.

\begin{table}[!htbp]
\caption{Average prediction error $\epsilon$ in days for fetal age prediction. Training and validation were performed on the distal hemispheres, while testing was performed on the distal and proximal hemispheres. Prediction errors $\epsilon_O$, $\epsilon_H$, and $\epsilon_{SR}$ are reported as mean $\pm$ standard deviation. Paired differences are reported as mean [95\% fetus-level cluster-bootstrap confidence interval], accounting for fetuses scanned at multiple gestational ages. Statistics were calculated over 263 scans from 252 fetuses. Statistical significance was assessed using a two-sided Wilcoxon signed-rank test, with differences from multiple scans of the same fetus first averaged so that each fetus contributed one independent observation. Asterisks mark mean paired differences significantly different from zero according to the Wilcoxon test after Bonferroni correction ($p<\frac{0.01}{4}=0.0025$).}
\label{tab:exp:ageprednum}
\centering
\resizebox{\linewidth}{!}{%
\begin{tabular}{@{}l|ccc|cc@{}}
\toprule
Hemisphere & $\epsilon_o$& $\epsilon_H$ & $\epsilon_{SR}$ & $\epsilon_O - \epsilon_{SR}$ & $\epsilon_H - \epsilon_{SR}$ \\
\midrule
Distal & $4.70 \pm 3.96$ & $4.27 \pm 3.24$ & $\mathbf{3.96 \pm 2.89}$ & $0.75\,[0.27, 1.22]$ & $0.31\,[-0.00, 0.64]$\\
Proximal & $12.66 \pm 8.76$ & $9.44 \pm 7.81$ & $\mathbf{7.60 \pm 5.73}$ & $5.08\,[4.00, 6.15]^*$ & $1.87\,[0.76, 2.96]^*$\\
\bottomrule
\end{tabular}
}
\end{table}

\begin{figure}[!htbp]
\centering
\includegraphics[width=\linewidth]{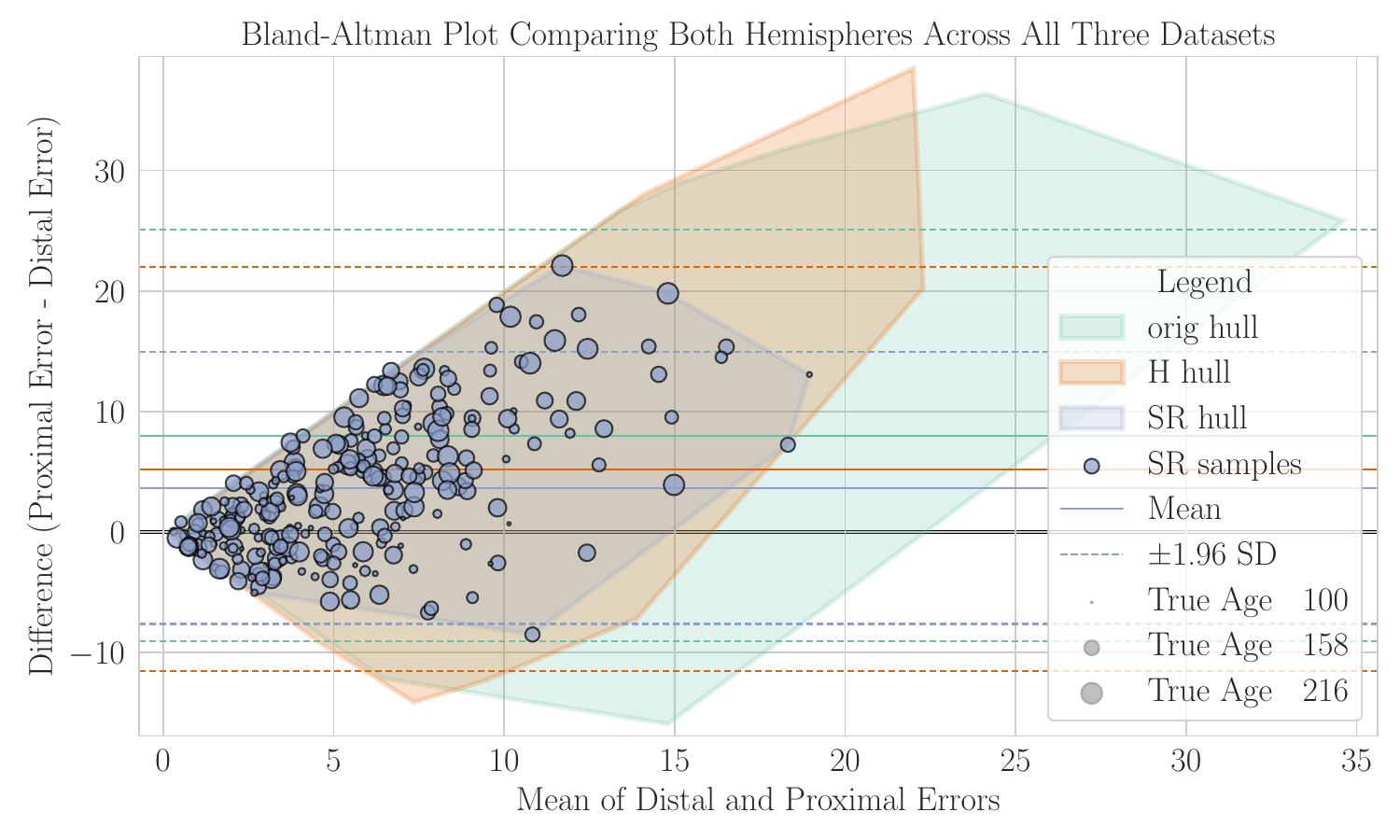}
\caption{Bland--Altman plots for age prediction on both hemispheres across the three datasets: the original images (orig), the images with Hughes shadow reduction (H), and the images with \proposed shadow reduction (SR). For readability, the first two datasets show only the convex hull, mean, and standard deviation. The full results are provided in \Cref{apdx:A:additional_age_pred}.}
\label{fig:results:bland_altman}
\end{figure}

\begin{figure}[!htbp]
    \centering
    \includegraphics[width=\linewidth]{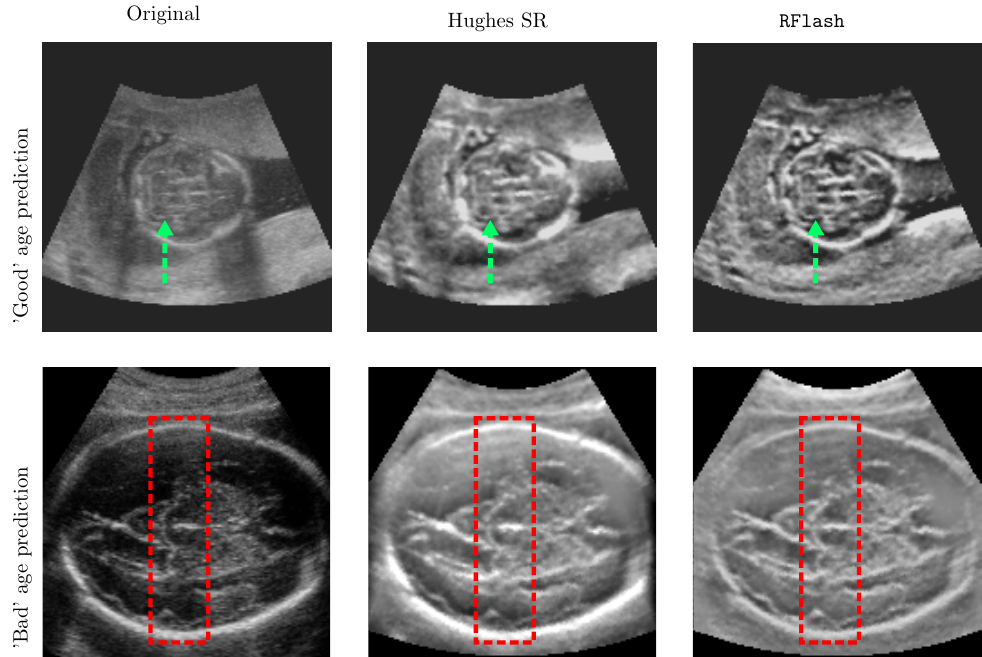}
    \caption{Representative cases from the age-prediction experiment. Top row: a case where shadow reduction substantially increased contrast in the proximal hemisphere. The green arrow marks the cerebellum, which is visibly better resolved after shadow reduction. Bottom row: a case where a residual grey-value gradient across the hemispheres persists (red box) in the original and Hughes-reduced images but is largely removed by \proposed.}
    \label{fig:results:age_pred}
\end{figure}

The individual samples showed considerable spread, reflecting differences in image quality.
\Cref{fig:results:bland_altman} shows the Bland--Altman plots for the original images, Hughes shadow reduction, and \proposed, plotting the difference between proximal and distal prediction errors against the mean error.
As expected, the distal hemisphere was more predictive on average, while \proposed improved predictability in both hemispheres with a stronger effect in the proximal hemisphere.
For readability, individual samples are shown only for \proposed, while the other methods are summarised by their convex hulls. The full distributions are provided in \Cref{apdx:A:additional_age_pred}.
Representative cases where proximal-hemisphere age prediction performed well and less well are shown in \Cref{fig:results:age_pred}.

\subsection{Shadow Confidence Maps}
\label{subsec:results:shadow_confidence_masks}

\begin{figure}[!htbp]
\centering
\setlength{\tabcolsep}{8pt}
\renewcommand{\arraystretch}{1.05}
\begin{tabular}{@{}cc@{}}
\scriptsize Meng & \scriptsize Ours\\
\includegraphics[width=0.38\textwidth]{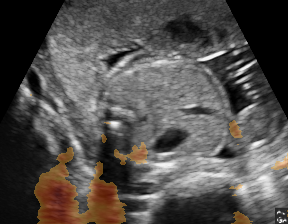} &
\includegraphics[width=0.38\textwidth]{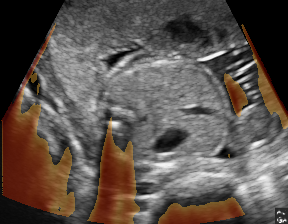}\\
\includegraphics[width=0.38\textwidth]{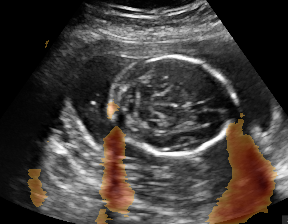} &
\includegraphics[width=0.38\textwidth]{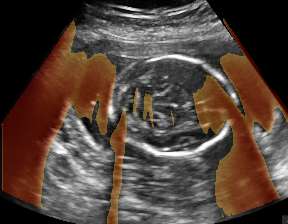}\\
\end{tabular}
\caption{Overlay comparison of the shadow confidence maps for two examples. The columns compare the overlays obtained from the method of Qingjie Meng et al. \cite{meng2019weakly} and our method.}
\label{fig:results:shadow_maps:overlay-comparison-meng}
\end{figure}

We evaluated the physical plausibility of the predicted shadow confidence maps using established confidence-map baselines.
The visual comparison in \Cref{fig:results:shadow_maps:overlay-comparison-meng} suggests that the maps from our physics-informed model produce more realistic shadow segmentations.

Compared with the physics-informed shadow maps from \citet{wysocki24ultranerf}, random forests using our shadow maps performed better in all metrics except precision, as shown in \Cref{tab:results:shadow_conf:numerical}.
\Cref{fig:results:conf_mask:visual_yesil} shows representative random-forest predictions.
Although the predictions remain imperfect, those based on our shadow confidence maps are visually more physically plausible because they segment shadows until the lower edge of the frame.
Analysis of SHAP \citep{lundberg17unified} feature-importance maps further indicates that the maps from \proposed were more informative to the random forest than those from \citet{yesilkaynak2024ultrasound}, as shown in \Cref{fig:results:shadow_conf:feature-maps}.
The colour maps are normalised column-wise to support comparison within each model configuration.

\begin{table}[!htbp]
\centering
\small
\setlength{\tabcolsep}{6pt}
\caption{Performance summary reported to three decimal places, averaged over 100 random forests. The baseline row used the ultrasound image as input, the second row used the shadow maps from Yesilkaynak \cite{yesilkaynak2024ultrasound}, and the third row used the shadow maps from our method. The best value in each column is highlighted in bold.}
\begin{tabular}{@{}lllll@{}}
\toprule
Name & Accuracy $\uparrow$ & Precision $\uparrow$ & Dice $\uparrow$ & Hausdorff dist. $\downarrow$\\
\midrule
Baseline
  & $0.912 \pm 0.006$ & $0.585 \pm 0.001$ & $0.532 \pm 0.003$ & $8.196 \pm 0.034$\\
Yesilkaynak & $0.922 \pm 0.006$ & $\mathbf{0.699 \pm 0.002}$ & $0.519 \pm 0.005$ & $5.404 \pm 0.188$\\
Ours
  & $\mathbf{0.930 \pm 0.005}$ & $0.693 \pm 0.001$ & $\mathbf{0.588 \pm 0.004}$ & $\mathbf{5.104 \pm 0.107}$\\
\bottomrule
\end{tabular}
\label{tab:results:shadow_conf:numerical}
\end{table}

\begin{figure}[!htbp]
\centering
\setlength{\tabcolsep}{2pt}
\renewcommand{\arraystretch}{1.05}
\begin{tabular}{@{}cccccc@{}}
\scriptsize Original & \scriptsize Baseline & \scriptsize Yesil. pred. & \scriptsize  Our prediction& \scriptsize Yesil. conf.map & \scriptsize Our confidence\\
\includegraphics[width=0.155\textwidth]{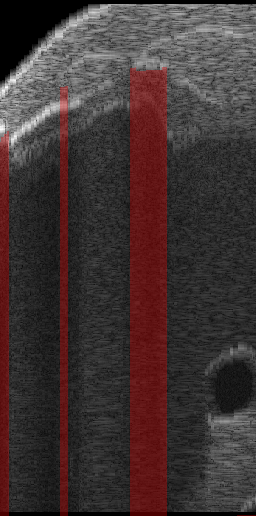} &
\includegraphics[width=0.155\textwidth]{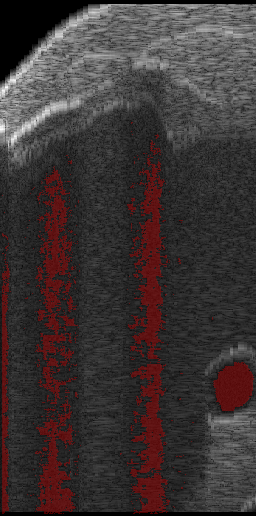} &
\includegraphics[width=0.155\textwidth]{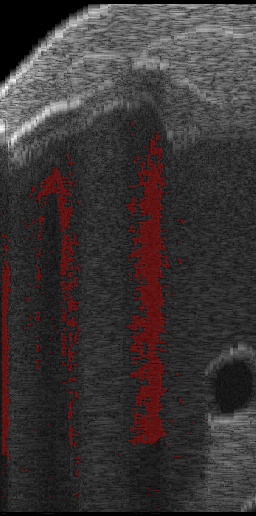} &
\includegraphics[width=0.155\textwidth]{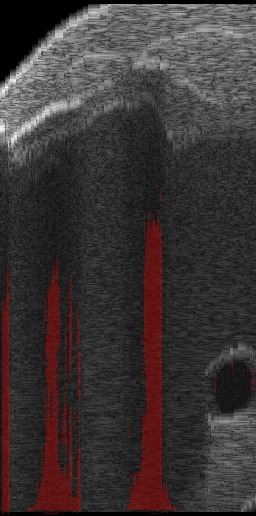} &
\includegraphics[width=0.155\textwidth]{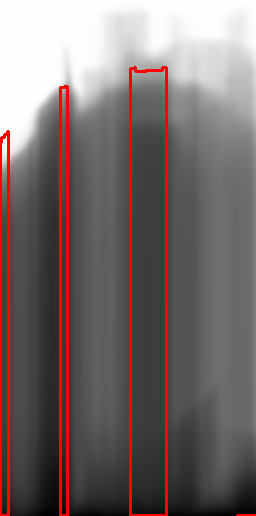} &
\includegraphics[width=0.155\textwidth]{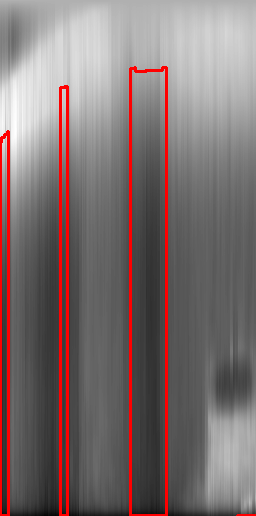} \\
\includegraphics[width=0.155\textwidth]{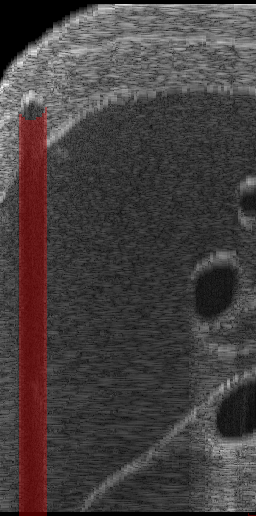} &
\includegraphics[width=0.155\textwidth]{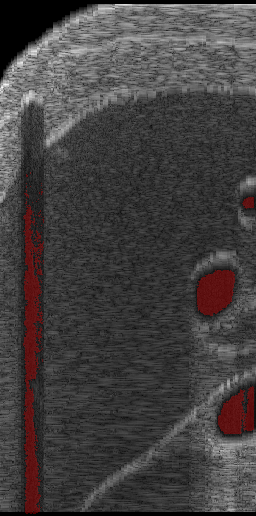} &
\includegraphics[width=0.155\textwidth]{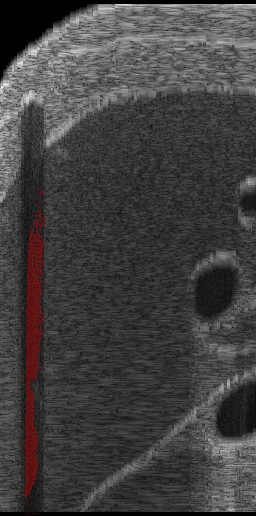} &
\includegraphics[width=0.155\textwidth]{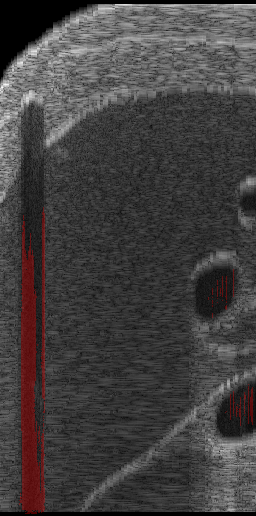} &
\includegraphics[width=0.155\textwidth]{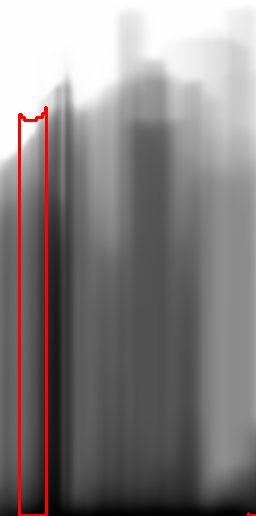} &
\includegraphics[width=0.155\textwidth]{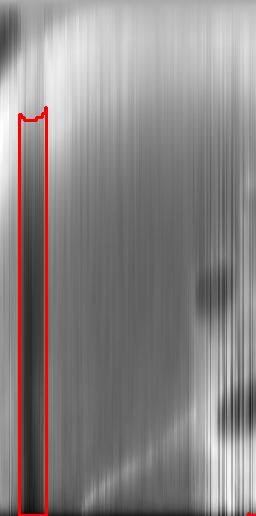} \\
\includegraphics[width=0.155\textwidth]{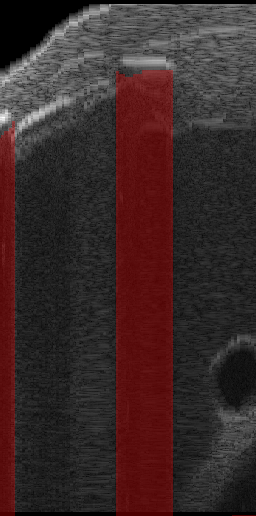} &
\includegraphics[width=0.155\textwidth]{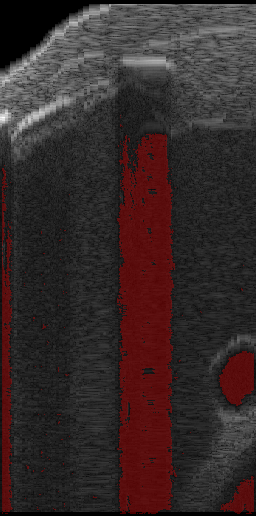} &
\includegraphics[width=0.155\textwidth]{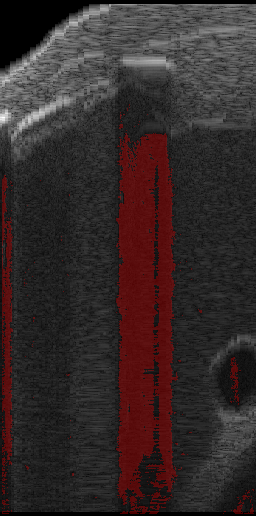} &
\includegraphics[width=0.155\textwidth]{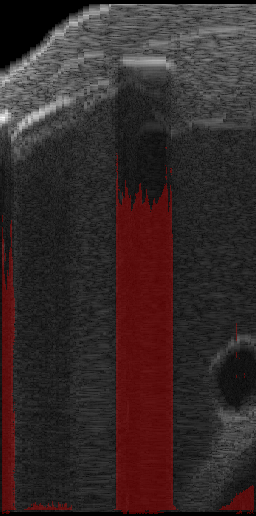} &
\includegraphics[width=0.155\textwidth]{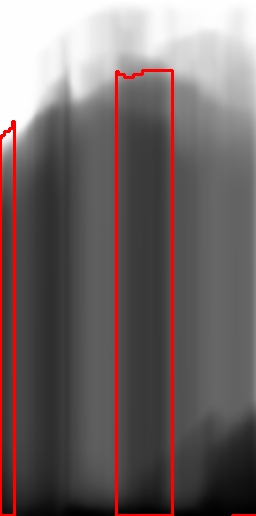} &
\includegraphics[width=0.155\textwidth]{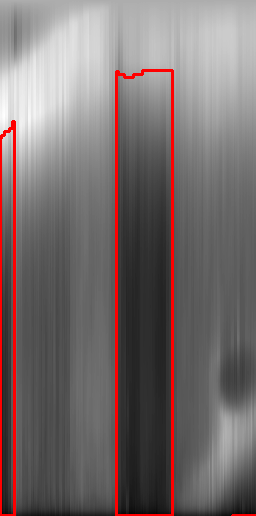} \\
\end{tabular}
\caption{Visual comparison for three selected ultrasound simulations of a liver phantom. The first column shows the ultrasound images with corresponding bone-shadow maps: the first row uses the ground-truth map, while the other rows use random-forest predictions. Yesil. refers to the methods introduced by \citep{yesilkaynak2024ultrasound}. The last two columns show confidence maps with the ground-truth outlines.}
\label{fig:results:conf_mask:visual_yesil}
\end{figure}

\begin{figure}[!htbp]
\centering
\newcommand{\explainabilitycell}[2]{%
\begin{minipage}{0.3\textwidth}
\centering
\includegraphics[height=0.66\textwidth]{figures/#1}\hfill
\includegraphics[height=0.66\textwidth]{figures/#2}%
\end{minipage}%
}
\newcommand{\emptyexplainabilitycell}{\makebox[0.30\textwidth]{}}
\newcommand{\explainabilitycolorbarcell}[1]{%
\begin{minipage}{0.3\textwidth}
\centering
\includegraphics[width=\linewidth]{figures/#1}%
\end{minipage}%
}
\newcommand{\explainabilityrowlabel}[1]{%
\rotatebox[origin=c]{90}{\makebox[0.2\textwidth][c]{\scriptsize #1}}%
}
\setlength{\tabcolsep}{3pt}
\begin{tabular}{@{}c ccc@{}}
& \scriptsize Baseline & \scriptsize Yesil. conf. map & \scriptsize Ours\\
\explainabilityrowlabel{SHAP maps} &
\explainabilitycell{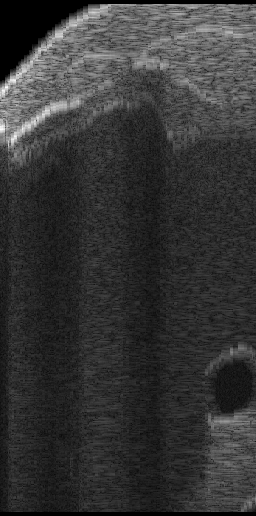}{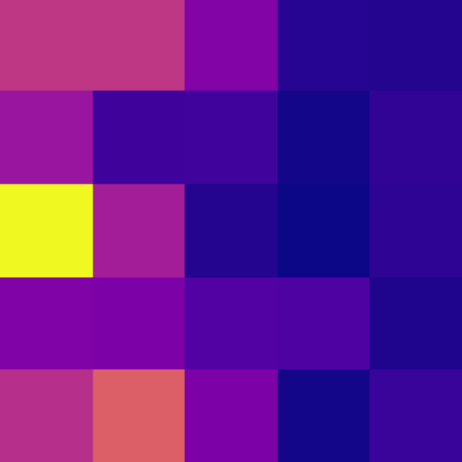} &
\explainabilitycell{us.png}{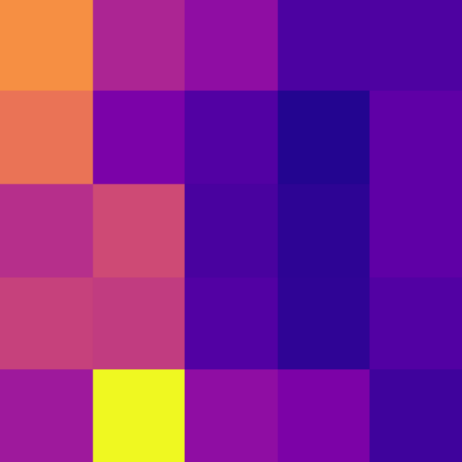} &
\explainabilitycell{us.png}{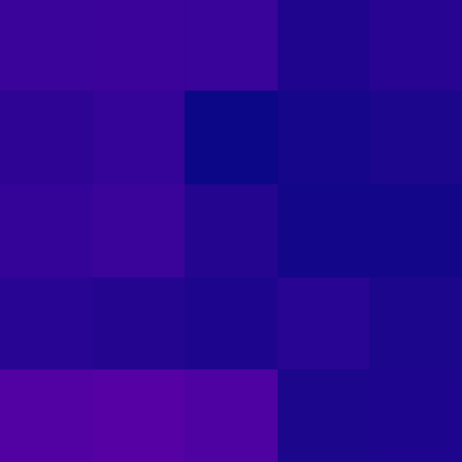}\\
&
\explainabilitycolorbarcell{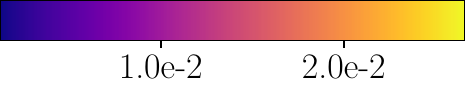} &
\explainabilitycolorbarcell{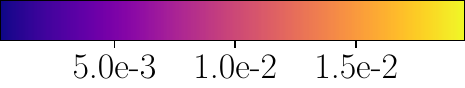} &
\explainabilitycolorbarcell{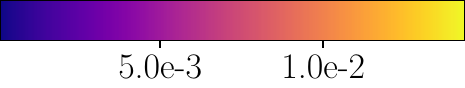}\\
\explainabilityrowlabel{Feature maps} &
\emptyexplainabilitycell &
\explainabilitycell{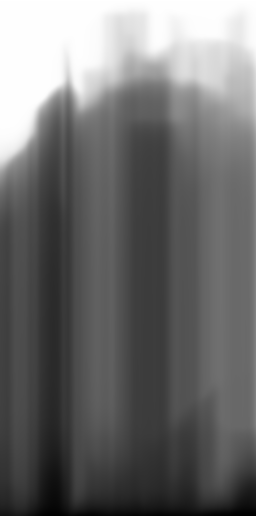}{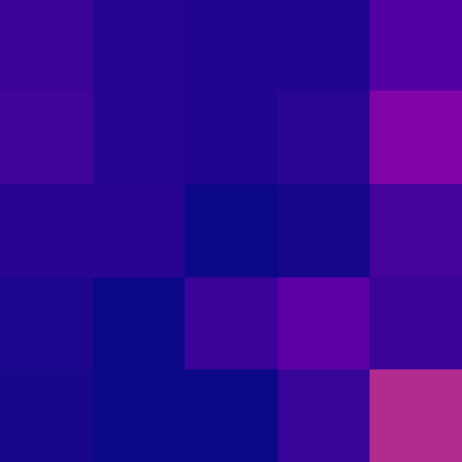} &
\explainabilitycell{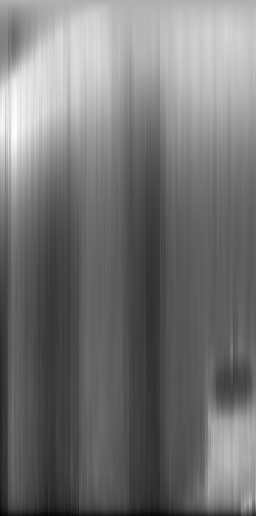}{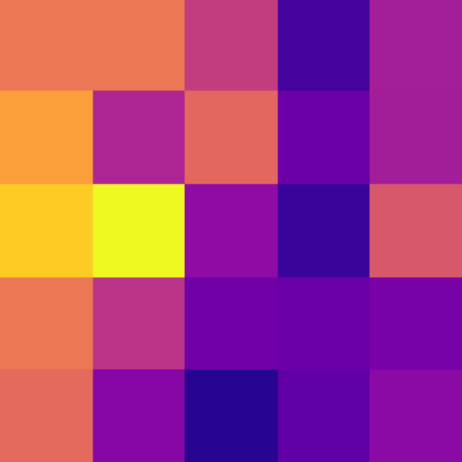}\\
\end{tabular}
\caption{Explainability maps for a single random forest computed from 1000 randomly selected paths using TreeExplainer. The columns correspond to the baseline model using the ultrasound image only, the model using the Yesilkaynak confidence map \cite{yesilkaynak2024ultrasound}, and our model. The first row shows SHAP feature-importance maps for the ultrasound images, the middle row shows the corresponding colour bars, and the final row shows the feature maps when available. The feature maps show the average importance of the $5\times5$-pixel patch around the classified centre pixel. The colours are matched within each random tree, that is, within each column. Higher SHAP magnitude on the \proposed confidence map (right) indicates that the random forest relied on it more than on the baseline map.}
\label{fig:results:shadow_conf:feature-maps}
\end{figure}

\section{Discussion}
\label{sec:discussion}

We presented \proposed, an iterative shadow-reduction method for 2D and 3D ultrasound images.
The method separates shadow reduction into two steps.
First, the input image is decomposed into \textit{physics-informed parameter maps} using a differentiable ultrasound simulation.
Second, these maps are used to \textit{render a shadow-reduced image}.
\proposed follows the radiance-field view of image formation \citep{mildenhall21nerf}, but uses a fully explicit representation instead of an MLP.
This representation is combined with a physics-informed renderer that models the path of beamformed sound through tissue and estimates attenuation and scatter intensity.
After decomposition, shadow-reduced images are rendered by reducing the dependence of each depth on tissue closer to the original transducer position, which can be interpreted as virtually moving the probe through the tissue.

This formulation avoids several limitations of previous shadow-reduction methods.
Classical approaches often rely on strong assumptions about the relation between attenuation and scattering \citep{hughes97automatic}, while frequency-based techniques require specialised hardware or signal information \citep{treece05ultrasound}.
In contrast, \proposed is applied as a post-processing method and does not depend on vendor-specific scanner access.
This makes it applicable across different ultrasound geometries, including 2D and 3D acquisitions. As it is trained on individual images, there is no knowledge transfer between patients, eliminating the risk of hallucinations.
Evaluation across these different probe geometries and acquisition types supports the robustness of \proposed to variation in the input ultrasound data.

The experiments show that \proposed can reduce shadows visually and quantitatively across different ultrasound image types.
On real data, it outperformed the Hughes shadow-reduction baseline \citep{hughes97automatic} in the proposed shadow metric and produced more uniform contrast.
In fetal brain ultrasound, the method increased the visibility of structures in the proximal hemisphere, which is typically more affected by acoustic shadows.
The downstream age-prediction experiment further suggests that the recovered information is clinically relevant: for a model trained on the distal hemisphere, shadow reduction improved prediction from the proximal hemisphere compared with both the original images and the baseline.

We also evaluated the physical consistency of the derived shadow confidence maps.
Because these maps are computed directly from the attenuation-related decomposition, they provide an interpretable estimate of shadow formation.
Compared with related confidence maps from \citet{wysocki24ultranerf}, our maps produced visually and quantitatively stronger guidance for random-forest shadow segmentation.
The SHAP analysis further indicated that the random forest assigned more importance to our shadow maps, suggesting that they contain more task-relevant shadow information.

As the method is trained on an individual-image basis, it can only recover structures that are still present in the image. 
Although this allows for more post-processing, it limits the structures that can be recovered. 
Applying the method to channel or RF data, which contain more information, may increase its clinical relevance, but reduce its applicability scanners which allow for RF data access. 
Further it relies on accurate geometry estimation. Quite often, the scanner geometry can be found in the scanners spec sheets. If this is not the case, geometry estimation is needed with is a non-trivial task, as seen for the abdominal dataset.
As an iterative method, \proposed is slower than closed-form solutions. Our code has not been optimised for speed, and runtime might be shortened with a hardware-optimised implementation.

\section*{Acknowledgements}
\label{sec:acknowledgements}

The first author was supported by a Clarendon Fund scholarship from the
University of Oxford. The work of Ana Namburete was supported by the Gates
Foundation. We thank the authors of \citet{meng18automatic} for providing us
with the shadow maps of their fetal head scans.

\section*{Declaration of Interests}
\label{sec:declaration_of_interests}

The authors declare that they have no known competing financial interests or personal relationships that could have appeared to influence the work reported in this paper. A priority application has been filed at the UK Intellectual Property Office (UKIPO) for the method described in this paper.

\section*{Declaration of generative AI and AI-assisted technologies in the manuscript preparation process.}
\label{sec:declaration_ai}

During the preparation of this work, the authors used GitHub Copilot (with ChatGPT 5.4 backend) and OpenAI's Codex (ChatGPT 5.5 and 5.6) to support code snipplets, generate figures from code-generated source files, and assist with the writing process. Generative AI was used exclusively to support implementation and improve the presentation of ideas, methods, and results. It was not used to generate the ideas, innovations, results, conclusions, or structure of the presented work. After using these tools, the authors reviewed and edited the content as needed and take full responsibility for the content of the published article.

\bibliographystyle{elsarticle-num-names}
\bibliography{references}

\begin{thebibliography}{59}
\expandafter\ifx\csname natexlab\endcsname\relax\def\natexlab#1{#1}\fi
\providecommand{\url}[1]{\texttt{#1}}
\providecommand{\href}[2]{#2}
\providecommand{\path}[1]{#1}
\providecommand{\DOIprefix}{doi:}
\providecommand{\ArXivprefix}{arXiv:}
\providecommand{\URLprefix}{URL: }
\providecommand{\Pubmedprefix}{pmid:}
\providecommand{\doi}[1]{\href{http://dx.doi.org/#1}{\path{#1}}}
\providecommand{\Pubmed}[1]{\href{pmid:#1}{\path{#1}}}
\providecommand{\bibinfo}[2]{#2}
\ifx\xfnm\relax \def\xfnm[#1]{\unskip,\space#1}\fi
\bibitem[{Pogledic et~al.(2024)Pogledic, Mankad, Severino, Lerman-Sagie, Jakab,
  Hadi, Jansen, Bahi-Buisson, Di~Donato, Oegema, Mitter, Capo, Whitehead,
  Haldipur, Mancini, Huisman, Righini, Dobyns, Barkovich, Jovanov~Milosevic,
  Kasprian, and Lequin}]{pogledic24prenatal}
\bibinfo{author}{I.~Pogledic}, \bibinfo{author}{K.~Mankad},
  \bibinfo{author}{M.~Severino}, \bibinfo{author}{T.~Lerman-Sagie},
  \bibinfo{author}{A.~Jakab}, \bibinfo{author}{E.~Hadi}, \bibinfo{author}{A.~C.
  Jansen}, \bibinfo{author}{N.~Bahi-Buisson}, \bibinfo{author}{N.~Di~Donato},
  \bibinfo{author}{R.~Oegema}, \bibinfo{author}{C.~Mitter},
  \bibinfo{author}{I.~Capo}, \bibinfo{author}{M.~T. Whitehead},
  \bibinfo{author}{P.~Haldipur}, \bibinfo{author}{G.~Mancini},
  \bibinfo{author}{T.~A. G.~M. Huisman}, \bibinfo{author}{A.~Righini},
  \bibinfo{author}{B.~Dobyns}, \bibinfo{author}{J.~A. Barkovich},
  \bibinfo{author}{N.~Jovanov~Milosevic}, \bibinfo{author}{G.~Kasprian},
  \bibinfo{author}{M.~Lequin},
\newblock \bibinfo{title}{{Prenatal assessment of brain malformations on
  neuroimaging: an expert panel review}},
\newblock \bibinfo{journal}{Brain}  (\bibinfo{year}{2024})
  \bibinfo{pages}{awae253}. \DOIprefix\doi{10.1093/brain/awae253}.
\bibitem[{Nelson et~al.(2000)Nelson, Pretorius, Hull, Riccabona, Sklansky, and
  James}]{nelson003dultrasound}
\bibinfo{author}{T.~R. Nelson}, \bibinfo{author}{D.~H. Pretorius},
  \bibinfo{author}{A.~Hull}, \bibinfo{author}{M.~Riccabona},
  \bibinfo{author}{M.~S. Sklansky}, \bibinfo{author}{G.~James},
\newblock \bibinfo{title}{Sources and impact of artifacts on clinical
  three-dimensional ultrasound imaging},
\newblock \bibinfo{journal}{Ultrasound in Obstetrics \& Gynecology}
  \bibinfo{volume}{16} (\bibinfo{year}{2000}) \bibinfo{pages}{374--383}.
  \DOIprefix\doi{10.1046/j.1469-0705.2000.00180.x}.
\bibitem[{Malinger et~al.(2020)Malinger, Paladini, Haratz, Monteagudo, Pilu,
  and Timor-Tritsch}]{malinger20ultrasound}
\bibinfo{author}{G.~Malinger}, \bibinfo{author}{D.~Paladini},
  \bibinfo{author}{K.~K. Haratz}, \bibinfo{author}{A.~Monteagudo},
  \bibinfo{author}{G.~L. Pilu}, \bibinfo{author}{I.~E. Timor-Tritsch},
\newblock \bibinfo{title}{Isuog practice guidelines (updated): sonographic
  examination of the fetal central nervous system. part 1: performance of
  screening examination and indications for targeted neurosonography},
\newblock \bibinfo{journal}{Ultrasound in Obstetrics \& Gynecology}
  \bibinfo{volume}{56} (\bibinfo{year}{2020}) \bibinfo{pages}{476--484}.
  \DOIprefix\doi{10.1002/uog.22145}.
\bibitem[{Paladini et~al.(2021)Paladini, Malinger, Birnbaum, Monteagudo, Pilu,
  Salomon, and Timor-Tritsch}]{paladini21ultrasound}
\bibinfo{author}{D.~Paladini}, \bibinfo{author}{G.~Malinger},
  \bibinfo{author}{R.~Birnbaum}, \bibinfo{author}{A.~Monteagudo},
  \bibinfo{author}{G.~Pilu}, \bibinfo{author}{L.~J. Salomon},
  \bibinfo{author}{I.~E. Timor-Tritsch},
\newblock \bibinfo{title}{Isuog practice guidelines (updated): sonographic
  examination of the fetal central nervous system. part 2: performance of
  targeted neurosonography},
\newblock \bibinfo{journal}{Ultrasound in Obstetrics \& Gynecology}
  \bibinfo{volume}{57} (\bibinfo{year}{2021}) \bibinfo{pages}{661--671}.
  \DOIprefix\doi{10.1002/uog.23616}.
\bibitem[{Namburete et~al.(2023)Namburete, Papież, Fernandes, Wyburd, Hesse,
  Moser, Ismail, Gunier, Squier, Ohuma, Carvalho, Jaffer, Gravett, Wu, Lambert,
  Winsey, Restrepo-Méndez, Bertino, Purwar, Barros, Stein, Noble, Molnár,
  Jenkinson, Bhutta, Papageorghiou, Villar, and Kennedy}]{namburete23normative}
\bibinfo{author}{A.~I.~L. Namburete}, \bibinfo{author}{B.~W. Papież},
  \bibinfo{author}{M.~Fernandes}, \bibinfo{author}{M.~K. Wyburd},
  \bibinfo{author}{L.~S. Hesse}, \bibinfo{author}{F.~A. Moser},
  \bibinfo{author}{L.~C. Ismail}, \bibinfo{author}{R.~B. Gunier},
  \bibinfo{author}{W.~Squier}, \bibinfo{author}{E.~O. Ohuma},
  \bibinfo{author}{M.~Carvalho}, \bibinfo{author}{Y.~Jaffer},
  \bibinfo{author}{M.~Gravett}, \bibinfo{author}{Q.~Wu},
  \bibinfo{author}{A.~Lambert}, \bibinfo{author}{A.~Winsey},
  \bibinfo{author}{M.~C. Restrepo-Méndez}, \bibinfo{author}{E.~Bertino},
  \bibinfo{author}{M.~Purwar}, \bibinfo{author}{F.~C. Barros},
  \bibinfo{author}{A.~Stein}, \bibinfo{author}{J.~A. Noble},
  \bibinfo{author}{Z.~Molnár}, \bibinfo{author}{M.~Jenkinson},
  \bibinfo{author}{Z.~A. Bhutta}, \bibinfo{author}{A.~T. Papageorghiou},
  \bibinfo{author}{J.~Villar}, \bibinfo{author}{S.~H. Kennedy},
\newblock \bibinfo{title}{Normative spatiotemporal fetal brain maturation with
  satisfactory development at 2 years},
\newblock \bibinfo{journal}{Nature} \bibinfo{volume}{623}
  (\bibinfo{year}{2023}) \bibinfo{pages}{106--114}.
  \DOIprefix\doi{10.1038/s41586-023-06630-3}.
\bibitem[{Xu et~al.(2022)Xu, Sanford, Turkbey, Xu, Wood, and
  Yan}]{xu22shadowconsistent}
\bibinfo{author}{X.~Xu}, \bibinfo{author}{T.~Sanford},
  \bibinfo{author}{B.~Turkbey}, \bibinfo{author}{S.~Xu}, \bibinfo{author}{B.~J.
  Wood}, \bibinfo{author}{P.~Yan},
\newblock \bibinfo{title}{Shadow-consistent semi-supervised learning for
  prostate ultrasound segmentation},
\newblock \bibinfo{journal}{IEEE Transactions on Medical Imaging}
  \bibinfo{volume}{41} (\bibinfo{year}{2022}) \bibinfo{pages}{1331--1345}.
  \DOIprefix\doi{10.1109/TMI.2021.3139999}.
\bibitem[{Hacihaliloglu(2017)}]{hacihaliloglu17enhancement}
\bibinfo{author}{I.~Hacihaliloglu},
\newblock \bibinfo{title}{Enhancement of bone shadow region using local
  phase-based ultrasound transmission maps},
\newblock \bibinfo{journal}{International Journal of Computer Assisted
  Radiology and Surgery} \bibinfo{volume}{12} (\bibinfo{year}{2017})
  \bibinfo{pages}{951--960}. \DOIprefix\doi{10.1007/s11548-017-1556-y}.
\bibitem[{Aldrich(2007)}]{aldrich07basic}
\bibinfo{author}{J.~E. Aldrich},
\newblock \bibinfo{title}{Basic physics of ultrasound imaging},
\newblock \bibinfo{journal}{Critical care medicine} \bibinfo{volume}{35}
  (\bibinfo{year}{2007}) \bibinfo{pages}{S131--S137}.
\bibitem[{Hughes and Duck(1997)}]{hughes97automatic}
\bibinfo{author}{D.~I. Hughes}, \bibinfo{author}{F.~A. Duck},
\newblock \bibinfo{title}{Automatic attenuation compensation for ultrasonic
  imaging},
\newblock \bibinfo{journal}{Ultrasound in Medicine \& Biology}
  \bibinfo{volume}{23} (\bibinfo{year}{1997}) \bibinfo{pages}{651--664}.
  \DOIprefix\doi{10.1016/S0301-5629(97)00002-1}.
\bibitem[{Knipp et~al.(1997)Knipp, Zagzebski, Wilson, Dong, and
  Madsen}]{knipp97attenuation}
\bibinfo{author}{B.~Knipp}, \bibinfo{author}{J.~Zagzebski},
  \bibinfo{author}{T.~Wilson}, \bibinfo{author}{F.~Dong},
  \bibinfo{author}{E.~Madsen},
\newblock \bibinfo{title}{Attenuation and backscatter estimation using video
  signal analysis applied to b-mode images},
\newblock \bibinfo{journal}{Ultrasonic Imaging} \bibinfo{volume}{19}
  (\bibinfo{year}{1997}) \bibinfo{pages}{221--233}.
  \DOIprefix\doi{10.1177/016173469701900305}, \bibinfo{note}{pMID: 9447670}.
\bibitem[{Treece et~al.(2005)Treece, Prager, and Gee}]{treece05ultrasound}
\bibinfo{author}{G.~Treece}, \bibinfo{author}{R.~Prager},
  \bibinfo{author}{A.~Gee},
\newblock \bibinfo{title}{Ultrasound attenuation measurement in the presence of
  scatterer variation for reduction of shadowing and enhancement},
\newblock \bibinfo{journal}{IEEE Transactions on Ultrasonics, Ferroelectrics,
  and Frequency Control} \bibinfo{volume}{52} (\bibinfo{year}{2005})
  \bibinfo{pages}{2346--2360}. \DOIprefix\doi{10.1109/TUFFC.2005.1563279}.
\bibitem[{Yasutomi et~al.(2019)Yasutomi, Arakaki, and
  Hamamoto}]{yasutomi19shadow}
\bibinfo{author}{S.~Yasutomi}, \bibinfo{author}{T.~Arakaki},
  \bibinfo{author}{R.~Hamamoto}, \bibinfo{title}{Shadow detection for
  ultrasound images using unlabeled data and synthetic shadows},
  \bibinfo{year}{2019}. \URLprefix \url{https://arxiv.org/abs/1908.01439}.
  \DOIprefix\doi{10.48550/ARXIV.1908.01439}.
\bibitem[{Yasutomi et~al.(2021)Yasutomi, Arakaki, Matsuoka, Sakai, Komatsu,
  Shozu, Dozen, Machino, Asada, Kaneko, Sekizawa, Hamamoto, and
  Komatsu}]{yasutomi21shadow}
\bibinfo{author}{S.~Yasutomi}, \bibinfo{author}{T.~Arakaki},
  \bibinfo{author}{R.~Matsuoka}, \bibinfo{author}{A.~Sakai},
  \bibinfo{author}{R.~Komatsu}, \bibinfo{author}{K.~Shozu},
  \bibinfo{author}{A.~Dozen}, \bibinfo{author}{H.~Machino},
  \bibinfo{author}{K.~Asada}, \bibinfo{author}{S.~Kaneko},
  \bibinfo{author}{A.~Sekizawa}, \bibinfo{author}{R.~Hamamoto},
  \bibinfo{author}{M.~Komatsu},
\newblock \bibinfo{title}{Shadow estimation for ultrasound images using
  auto-encoding structures and synthetic shadows},
\newblock \bibinfo{journal}{Applied Sciences} \bibinfo{volume}{11}
  (\bibinfo{year}{2021}). \URLprefix
  \url{https://www.mdpi.com/2076-3417/11/3/1127}.
  \DOIprefix\doi{10.3390/app11031127}.
\bibitem[{Hewlett E.~Melton and David J.~Skorton(1983)}]{melton83rational}
\bibinfo{author}{J.~Hewlett E.~Melton}, \bibinfo{author}{J.~David J.~Skorton},
\newblock \bibinfo{title}{Rational gain compensation for attenuation in cardiac
  ultrasonography},
\newblock \bibinfo{journal}{Ultrasonic Imaging} \bibinfo{volume}{5}
  (\bibinfo{year}{1983}) \bibinfo{pages}{214--228}.
  \DOIprefix\doi{10.1177/016173468300500302}, \bibinfo{note}{pMID: 6685367}.
\bibitem[{Girard et~al.(2011)Girard, Strouthidis, Ethier, and
  Mari}]{girard11shadow}
\bibinfo{author}{M.~J.~A. Girard}, \bibinfo{author}{N.~G. Strouthidis},
  \bibinfo{author}{C.~R. Ethier}, \bibinfo{author}{J.~M. Mari},
\newblock \bibinfo{title}{{Shadow Removal and Contrast Enhancement in Optical
  Coherence Tomography Images of the Human Optic Nerve Head}},
\newblock \bibinfo{journal}{Investigative Ophthalmology \& Visual Science}
  \bibinfo{volume}{52} (\bibinfo{year}{2011}) \bibinfo{pages}{7738--7748}.
  \DOIprefix\doi{10.1167/iovs.10-6925}.
\bibitem[{Vermeer et~al.(2014)Vermeer, Mo, Weda, Lemij, and
  de~Boer}]{vermeer14depthResolved}
\bibinfo{author}{K.~A. Vermeer}, \bibinfo{author}{J.~Mo},
  \bibinfo{author}{J.~J.~A. Weda}, \bibinfo{author}{H.~G. Lemij},
  \bibinfo{author}{J.~F. de~Boer},
\newblock \bibinfo{title}{Depth-resolved model-based reconstruction of
  attenuation coefficients in optical coherence tomography},
\newblock \bibinfo{journal}{Biomed. Opt. Express} \bibinfo{volume}{5}
  (\bibinfo{year}{2014}) \bibinfo{pages}{322--337}.
  \DOIprefix\doi{10.1364/BOE.5.000322}.
\bibitem[{Kim and Varghese(2008)}]{kim08hybrid}
\bibinfo{author}{H.~Kim}, \bibinfo{author}{T.~Varghese},
\newblock \bibinfo{title}{Hybrid spectral domain method for attenuation slope
  estimation},
\newblock \bibinfo{journal}{Ultrasound in Medicine \& Biology}
  \bibinfo{volume}{34} (\bibinfo{year}{2008}) \bibinfo{pages}{1808--1819}.
  \DOIprefix\doi{10.1016/j.ultrasmedbio.2008.04.011}.
\bibitem[{Yu and Wang(2010)}]{yu10backscatter}
\bibinfo{author}{Y.~Yu}, \bibinfo{author}{J.~Wang},
\newblock \bibinfo{title}{Backscatter-contour-attenuation joint estimation
  model for attenuation compensation in ultrasound imagery},
\newblock \bibinfo{journal}{IEEE Transactions on Image Processing}
  \bibinfo{volume}{19} (\bibinfo{year}{2010}) \bibinfo{pages}{2725--2736}.
  \DOIprefix\doi{10.1109/TIP.2010.2050636}.
\bibitem[{Mildenhall et~al.(2021)Mildenhall, Srinivasan, Tancik, Barron,
  Ramamoorthi, and Ng}]{mildenhall21nerf}
\bibinfo{author}{B.~Mildenhall}, \bibinfo{author}{P.~P. Srinivasan},
  \bibinfo{author}{M.~Tancik}, \bibinfo{author}{J.~T. Barron},
  \bibinfo{author}{R.~Ramamoorthi}, \bibinfo{author}{R.~Ng},
\newblock \bibinfo{title}{Nerf: representing scenes as neural radiance fields
  for view synthesis},
\newblock \bibinfo{journal}{Communications of the ACM} \bibinfo{volume}{65}
  (\bibinfo{year}{2021}) \bibinfo{pages}{99--106}.
\bibitem[{Gao et~al.(2026)Gao, Gao, He, Lu, Xu, and Li}]{gao23nerf}
\bibinfo{author}{K.~Gao}, \bibinfo{author}{Y.~Gao}, \bibinfo{author}{H.~He},
  \bibinfo{author}{D.~Lu}, \bibinfo{author}{L.~Xu}, \bibinfo{author}{J.~Li},
\newblock \bibinfo{title}{Neural radiance fields in 3d vision: A comprehensive
  review},
\newblock \bibinfo{journal}{Computational Visual Media}  (\bibinfo{year}{2026})
  \bibinfo{pages}{1--55}. \DOIprefix\doi{10.26599/CVM.2026.9450535}.
\bibitem[{Yeung et~al.(2024)Yeung, Hesse, Aliasi, Haak, Xie, and
  Namburete}]{yeung24sensorless}
\bibinfo{author}{P.-H. Yeung}, \bibinfo{author}{L.~S. Hesse},
  \bibinfo{author}{M.~Aliasi}, \bibinfo{author}{M.~C. Haak},
  \bibinfo{author}{W.~Xie}, \bibinfo{author}{A.~I. Namburete},
\newblock \bibinfo{title}{Sensorless volumetric reconstruction of fetal brain
  freehand ultrasound scans with deep implicit representation},
\newblock \bibinfo{journal}{Medical Image Analysis} \bibinfo{volume}{94}
  (\bibinfo{year}{2024}) \bibinfo{pages}{103147}.
  \DOIprefix\doi{10.1016/j.media.2024.103147}.
\bibitem[{Wysocki et~al.(2024)Wysocki, Azampour, Eilers, Busam, Salehi, and
  Navab}]{wysocki24ultranerf}
\bibinfo{author}{M.~Wysocki}, \bibinfo{author}{M.~F. Azampour},
  \bibinfo{author}{C.~Eilers}, \bibinfo{author}{B.~Busam},
  \bibinfo{author}{M.~Salehi}, \bibinfo{author}{N.~Navab},
\newblock \bibinfo{title}{Ultra-nerf: Neural radiance fields for ultrasound
  imaging},
\newblock in: \bibinfo{editor}{I.~Oguz}, \bibinfo{editor}{J.~Noble},
  \bibinfo{editor}{X.~Li}, \bibinfo{editor}{M.~Styner},
  \bibinfo{editor}{C.~Baumgartner}, \bibinfo{editor}{M.~Rusu},
  \bibinfo{editor}{T.~Heinmann}, \bibinfo{editor}{D.~Kontos},
  \bibinfo{editor}{B.~Landman}, \bibinfo{editor}{B.~Dawant} (Eds.),
  \bibinfo{booktitle}{Medical Imaging with Deep Learning}, volume
  \bibinfo{volume}{227} of \textit{\bibinfo{series}{Proceedings of Machine
  Learning Research}}, \bibinfo{publisher}{PMLR}, \bibinfo{year}{2024}, pp.
  \bibinfo{pages}{382--401}. \URLprefix
  \url{https://proceedings.mlr.press/v227/wysocki24a.html}.
\bibitem[{Dagli et~al.(2024)Dagli, Hibi, Krishnan, and Tyrrell}]{dagli24nerfus}
\bibinfo{author}{R.~Dagli}, \bibinfo{author}{A.~Hibi},
  \bibinfo{author}{R.~Krishnan}, \bibinfo{author}{P.~N. Tyrrell},
\newblock \bibinfo{title}{Ne{RF}-{US}: Removing ultrasound imaging artifacts
  from neural radiance fields in the wild},
\newblock in: \bibinfo{editor}{K.~Deshpande}, \bibinfo{editor}{M.~Fiterau},
  \bibinfo{editor}{S.~Joshi}, \bibinfo{editor}{Z.~Lipton},
  \bibinfo{editor}{R.~Ranganath}, \bibinfo{editor}{I.~Urteaga} (Eds.),
  \bibinfo{booktitle}{Proceedings of the 9th Machine Learning for Healthcare
  Conference}, volume \bibinfo{volume}{252} of
  \textit{\bibinfo{series}{Proceedings of Machine Learning Research}},
  \bibinfo{publisher}{PMLR}, \bibinfo{year}{2024}. \URLprefix
  \url{https://proceedings.mlr.press/v252/dagli24a.html}.
\bibitem[{Gaits et~al.(2024)Gaits, Mellado, and Basarab}]{gaits24ultrasound}
\bibinfo{author}{F.~Gaits}, \bibinfo{author}{N.~Mellado},
  \bibinfo{author}{A.~Basarab},
\newblock \bibinfo{title}{{Ultrasound volume reconstruction from 2D Freehand
  acquisitions using neural implicit representations}},
\newblock in: \bibinfo{booktitle}{{21st IEEE International Symposium on
  Biomedical Imaging (ISBI 2024)}}, \bibinfo{organization}{{IEEE Signal
  Processing Society and IEEE Engineering in Medicine and Biology Society}},
  \bibinfo{address}{Ath{\`e}nes, Greece}, \bibinfo{year}{2024}, p.
  \bibinfo{pages}{{\`a} para{\^i}tre}. \URLprefix
  \url{https://hal.science/hal-04480668}.
\bibitem[{Hu et~al.(2024)Hu, Zou, Xiao, and Chen}]{hu24neural}
\bibinfo{author}{X.~Hu}, \bibinfo{author}{B.~Zou}, \bibinfo{author}{Z.~Xiao},
  \bibinfo{author}{Q.~Chen},
\newblock \bibinfo{title}{Neural radiance fields for ultrasound imaging with
  image-to-image refinements},
\newblock in: \bibinfo{booktitle}{2024 4th International Conference on Neural
  Networks, Information and Communication Engineering (NNICE)},
  \bibinfo{year}{2024}, pp. \bibinfo{pages}{355--358}.
  \DOIprefix\doi{10.1109/NNICE61279.2024.10498317}.
\bibitem[{Zhang et~al.(2025)Zhang, Zhao, and Ouyang}]{zhang25hfusnerf}
\bibinfo{author}{S.~Zhang}, \bibinfo{author}{C.~Zhao},
  \bibinfo{author}{B.~Ouyang},
\newblock \bibinfo{title}{Hfus-nerf: Hybrid representation for fast ultrasound
  reconstruction in robotic ultrasound system},
\newblock in: \bibinfo{booktitle}{2025 IEEE International Conference on
  Robotics and Automation (ICRA)}, \bibinfo{year}{2025}, pp.
  \bibinfo{pages}{4092--4098}. \DOIprefix\doi{10.1109/ICRA55743.2025.11128742}.
\bibitem[{Eid et~al.(2025)Eid, Yeung, Wyburd, Henriques, and
  Namburete}]{eid24rapidvol}
\bibinfo{author}{M.~C. Eid}, \bibinfo{author}{P.-H. Yeung},
  \bibinfo{author}{M.~K. Wyburd}, \bibinfo{author}{J.~F. Henriques},
  \bibinfo{author}{A.~I. Namburete},
\newblock \bibinfo{title}{Rapidvol: Rapid reconstruction of 3d ultrasound
  volumes from sensorless 2d scans},
\newblock in: \bibinfo{booktitle}{2025 IEEE 22nd International Symposium on
  Biomedical Imaging (ISBI)}, \bibinfo{year}{2025}, pp. \bibinfo{pages}{1--5}.
  \DOIprefix\doi{10.1109/ISBI60581.2025.10980994}.
\bibitem[{Guo et~al.(2024)Guo, Fang, and Fu}]{guo24ulrenerf}
\bibinfo{author}{Z.~Guo}, \bibinfo{author}{Z.~Fang}, \bibinfo{author}{Z.~Fu},
  \bibinfo{title}{Ulre-nerf: 3d ultrasound imaging through neural rendering
  with ultrasound reflection direction parameterization}, \bibinfo{year}{2024}.
  \URLprefix \url{https://arxiv.org/abs/2408.00860}.
  \href{http://arxiv.org/abs/2408.00860}{{\tt arXiv:2408.00860}}.
\bibitem[{Eid et~al.(2026)Eid, Namburete, and Henriques}]{eid25ultragauss}
\bibinfo{author}{M.~Eid}, \bibinfo{author}{A.~Namburete},
  \bibinfo{author}{J.~F. Henriques},
\newblock \bibinfo{title}{Ultragauss: Ultrafast gaussian reconstruction of 3d
  ultrasound volumes},
\newblock in: \bibinfo{editor}{C.~Vondrick}, \bibinfo{editor}{B.~Hariharan},
  \bibinfo{editor}{C.~Raffel}, \bibinfo{editor}{L.~Pinto},
  \bibinfo{editor}{D.~Yang}, \bibinfo{editor}{A.~Faust} (Eds.),
  \bibinfo{booktitle}{International Conference on Learning Representations},
  volume \bibinfo{volume}{2026}, \bibinfo{year}{2026}, pp.
  \bibinfo{pages}{93682--93716}. \URLprefix
  \url{https://proceedings.iclr.cc/paper_files/paper/2026/file/9739fdfbecb84b2cab3ba06f3ee5498b-Paper-Conference.pdf}.
\bibitem[{Chen et~al.(2022)Chen, Xu, Geiger, Yu, and Su}]{chen22tensorf}
\bibinfo{author}{A.~Chen}, \bibinfo{author}{Z.~Xu},
  \bibinfo{author}{A.~Geiger}, \bibinfo{author}{J.~Yu},
  \bibinfo{author}{H.~Su},
\newblock \bibinfo{title}{Tensorf: Tensorial radiance fields},
\newblock in: \bibinfo{booktitle}{Computer Vision -- ECCV 2022: 17th European
  Conference, Tel Aviv, Israel, October 23--27, 2022, Proceedings, Part XXXII},
  \bibinfo{publisher}{Springer-Verlag}, \bibinfo{address}{Berlin, Heidelberg},
  \bibinfo{year}{2022}, pp. \bibinfo{pages}{333--350}.
  \DOIprefix\doi{10.1007/978-3-031-19824-3\_20}.
\bibitem[{Chan et~al.(2022)Chan, Lin, Chan, Nagano, Pan, de~Mello, Gallo,
  Guibas, Tremblay, Khamis, Karras, and Wetzstein}]{chan22efficient}
\bibinfo{author}{E.~R. Chan}, \bibinfo{author}{C.~Z. Lin},
  \bibinfo{author}{M.~A. Chan}, \bibinfo{author}{K.~Nagano},
  \bibinfo{author}{B.~Pan}, \bibinfo{author}{S.~de~Mello},
  \bibinfo{author}{O.~Gallo}, \bibinfo{author}{L.~Guibas},
  \bibinfo{author}{J.~Tremblay}, \bibinfo{author}{S.~Khamis},
  \bibinfo{author}{T.~Karras}, \bibinfo{author}{G.~Wetzstein},
\newblock \bibinfo{title}{Efficient geometry-aware 3d generative adversarial
  networks},
\newblock in: \bibinfo{booktitle}{2022 IEEE/CVF Conference on Computer Vision
  and Pattern Recognition (CVPR)}, \bibinfo{year}{2022}, pp.
  \bibinfo{pages}{16102--16112}. \DOIprefix\doi{10.1109/CVPR52688.2022.01565}.
\bibitem[{Bamber and Dickinson(1980)}]{bamber80ultrasonic}
\bibinfo{author}{J.~C. Bamber}, \bibinfo{author}{R.~J. Dickinson},
\newblock \bibinfo{title}{Ultrasonic b-scanning: a computer simulation},
\newblock \bibinfo{journal}{Physics in Medicine \& Biology}
  \bibinfo{volume}{25} (\bibinfo{year}{1980}) \bibinfo{pages}{463}.
  \DOIprefix\doi{10.1088/0031-9155/25/3/006}.
\bibitem[{Meunier and Bertrand(1995)}]{meunier95ultrasonic}
\bibinfo{author}{J.~Meunier}, \bibinfo{author}{M.~Bertrand},
\newblock \bibinfo{title}{Ultrasonic texture motion analysis: theory and
  simulation},
\newblock \bibinfo{journal}{IEEE Transactions on Medical Imaging}
  \bibinfo{volume}{14} (\bibinfo{year}{1995}) \bibinfo{pages}{293--300}.
  \DOIprefix\doi{10.1109/42.387711}.
\bibitem[{B{\"u}rger et~al.(2008)B{\"u}rger, Abkai, and
  Hesser}]{burger08simulation}
\bibinfo{author}{B.~B{\"u}rger}, \bibinfo{author}{C.~Abkai},
  \bibinfo{author}{J.~Hesser},
\newblock \bibinfo{title}{Simulation of dynamic ultrasound based on ct models
  for medical education},
\newblock \bibinfo{journal}{Studies in health technology and informatics}
  \bibinfo{volume}{132} (\bibinfo{year}{2008}) \bibinfo{pages}{56}.
\bibitem[{Hergum et~al.(2009)Hergum, Langeland, Remme, and Torp}]{hergum09fast}
\bibinfo{author}{T.~Hergum}, \bibinfo{author}{S.~Langeland},
  \bibinfo{author}{E.~W. Remme}, \bibinfo{author}{H.~Torp},
\newblock \bibinfo{title}{Fast ultrasound imaging simulation in k-space},
\newblock \bibinfo{journal}{IEEE Transactions on Ultrasonics, Ferroelectrics,
  and Frequency Control} \bibinfo{volume}{56} (\bibinfo{year}{2009})
  \bibinfo{pages}{1159--1167}. \DOIprefix\doi{10.1109/TUFFC.2009.1158}.
\bibitem[{Marion and Vray(2009)}]{marion09toward}
\bibinfo{author}{A.~Marion}, \bibinfo{author}{D.~Vray},
\newblock \bibinfo{title}{Toward a real-time simulation of ultrasound image
  sequences based on a 3-d set of moving scatterers},
\newblock \bibinfo{journal}{IEEE Transactions on Ultrasonics, Ferroelectrics,
  and Frequency Control} \bibinfo{volume}{56} (\bibinfo{year}{2009})
  \bibinfo{pages}{2167--2179}. \DOIprefix\doi{10.1109/TUFFC.2009.1299}.
\bibitem[{Mattausch and Goksel(2018)}]{mattausch18imagebased}
\bibinfo{author}{O.~Mattausch}, \bibinfo{author}{O.~Goksel},
\newblock \bibinfo{title}{Image-based reconstruction of tissue scatterers using
  beam steering for ultrasound simulation},
\newblock \bibinfo{journal}{IEEE Transactions on Medical Imaging}
  \bibinfo{volume}{37} (\bibinfo{year}{2018}) \bibinfo{pages}{767--780}.
  \DOIprefix\doi{10.1109/TMI.2017.2770118}.
\bibitem[{Gjerald et~al.(2012)Gjerald, Brekken, Hergum, and
  D'hooge}]{gjerald12realtime}
\bibinfo{author}{S.~U. Gjerald}, \bibinfo{author}{R.~Brekken},
  \bibinfo{author}{T.~Hergum}, \bibinfo{author}{J.~D'hooge},
\newblock \bibinfo{title}{Real-time ultrasound simulation using the gpu},
\newblock \bibinfo{journal}{IEEE Transactions on Ultrasonics, Ferroelectrics,
  and Frequency Control} \bibinfo{volume}{59} (\bibinfo{year}{2012})
  \bibinfo{pages}{885--892}. \DOIprefix\doi{10.1109/TUFFC.2012.2273}.
\bibitem[{Wein et~al.(2008)Wein, Brunke, Khamene, Callstrom, and
  Navab}]{wein08automatic}
\bibinfo{author}{W.~Wein}, \bibinfo{author}{S.~Brunke},
  \bibinfo{author}{A.~Khamene}, \bibinfo{author}{M.~R. Callstrom},
  \bibinfo{author}{N.~Navab},
\newblock \bibinfo{title}{Automatic ct-ultrasound registration for diagnostic
  imaging and image-guided intervention},
\newblock \bibinfo{journal}{Medical Image Analysis} \bibinfo{volume}{12}
  (\bibinfo{year}{2008}) \bibinfo{pages}{577--585}.
  \DOIprefix\doi{10.1016/j.media.2008.06.006}, \bibinfo{note}{special issue on
  the 10th international conference on medical imaging and computer assisted
  intervention - MICCAI 2007}.
\bibitem[{Law et~al.(2011)Law, Ullrich, Knott, Kuhlen, and
  Weg}]{law11ultrasound}
\bibinfo{author}{Y.~C. Law}, \bibinfo{author}{S.~Ullrich},
  \bibinfo{author}{T.~Knott}, \bibinfo{author}{T.~Kuhlen},
  \bibinfo{author}{S.~Weg},
\newblock \bibinfo{title}{Ultrasound image simulation with gpu-based ray
  tracing},
\newblock \bibinfo{journal}{Virtuelle und Erweiterte Realit{\"a}t}
  \bibinfo{volume}{8} (\bibinfo{year}{2011}) \bibinfo{pages}{183--194}.
\bibitem[{Burger et~al.(2013)Burger, Bettinghausen, Radle, and
  Hesser}]{burger13realtime}
\bibinfo{author}{B.~Burger}, \bibinfo{author}{S.~Bettinghausen},
  \bibinfo{author}{M.~Radle}, \bibinfo{author}{J.~Hesser},
\newblock \bibinfo{title}{Real-time gpu-based ultrasound simulation using
  deformable mesh models},
\newblock \bibinfo{journal}{IEEE Transactions on Medical Imaging}
  \bibinfo{volume}{32} (\bibinfo{year}{2013}) \bibinfo{pages}{609--618}.
  \DOIprefix\doi{10.1109/TMI.2012.2234474}.
\bibitem[{Salehi et~al.(2015)Salehi, Ahmadi, Prevost, Navab, and
  Wein}]{salehi15patient}
\bibinfo{author}{M.~Salehi}, \bibinfo{author}{S.-A. Ahmadi},
  \bibinfo{author}{R.~Prevost}, \bibinfo{author}{N.~Navab},
  \bibinfo{author}{W.~Wein},
\newblock \bibinfo{title}{Patient-specific 3d ultrasound simulation based on
  convolutional ray-tracing and appearance optimization},
\newblock in: \bibinfo{editor}{N.~Navab}, \bibinfo{editor}{J.~Hornegger},
  \bibinfo{editor}{W.~M. Wells}, \bibinfo{editor}{A.~Frangi} (Eds.),
  \bibinfo{booktitle}{Medical Image Computing and Computer-Assisted
  Intervention -- MICCAI 2015}, \bibinfo{publisher}{Springer International
  Publishing}, \bibinfo{address}{Cham}, \bibinfo{year}{2015}, pp.
  \bibinfo{pages}{510--518}.
\bibitem[{Mattausch and Goksel(2016)}]{mattausch16monteCarlo}
\bibinfo{author}{O.~Mattausch}, \bibinfo{author}{O.~Goksel},
\newblock \bibinfo{title}{{Monte-Carlo Ray-Tracing for Realistic Interactive
  Ultrasound Simulation}},
\newblock in: \bibinfo{editor}{S.~Bruckner}, \bibinfo{editor}{B.~Preim},
  \bibinfo{editor}{A.~Vilanova}, \bibinfo{editor}{H.~Hauser},
  \bibinfo{editor}{A.~Hennemuth}, \bibinfo{editor}{A.~Lundervold} (Eds.),
  \bibinfo{booktitle}{Eurographics Workshop on Visual Computing for Biology and
  Medicine}, \bibinfo{publisher}{The Eurographics Association},
  \bibinfo{year}{2016}. \DOIprefix\doi{10.2312/vcbm.20161285}.
\bibitem[{Mattausch et~al.(2018)Mattausch, Makhinya, and
  Goksel}]{mattausch18realistic}
\bibinfo{author}{O.~Mattausch}, \bibinfo{author}{M.~Makhinya},
  \bibinfo{author}{O.~Goksel},
\newblock \bibinfo{title}{Realistic ultrasound simulation of complex surface
  models using interactive monte-carlo path tracing},
\newblock \bibinfo{journal}{Computer Graphics Forum} \bibinfo{volume}{37}
  (\bibinfo{year}{2018}) \bibinfo{pages}{202--213}.
  \DOIprefix\doi{10.1111/cgf.13260}.
\bibitem[{Duelmer et~al.(2025)Duelmer, Azampour, Wysocki, and
  Navab}]{duelmer2025ultraray}
\bibinfo{author}{F.~Duelmer}, \bibinfo{author}{M.~F. Azampour},
  \bibinfo{author}{M.~Wysocki}, \bibinfo{author}{N.~Navab},
\newblock \bibinfo{title}{Ultraray: Introducing full-path ray tracing in
  physics-based ultrasound simulation},
\newblock in: \bibinfo{booktitle}{International Conference on Medical Image
  Computing and Computer-Assisted Intervention},
  \bibinfo{organization}{Springer}, \bibinfo{year}{2025}, pp.
  \bibinfo{pages}{653--662}.
\bibitem[{Bertramo et~al.(2026)Bertramo, Duguey, and
  Gopalakrishnan}]{bertramo25diffus}
\bibinfo{author}{N.~Bertramo}, \bibinfo{author}{G.~Duguey},
  \bibinfo{author}{V.~Gopalakrishnan},
\newblock \bibinfo{title}{Diffus: Differentiable ultrasound rendering
  from volumetric imaging},
\newblock in: \bibinfo{editor}{D.~Ni}, \bibinfo{editor}{A.~Noble},
  \bibinfo{editor}{R.~Huang}, \bibinfo{editor}{W.~Xue} (Eds.),
  \bibinfo{booktitle}{Simplifying Medical Ultrasound},
  \bibinfo{publisher}{Springer Nature Switzerland}, \bibinfo{address}{Cham},
  \bibinfo{year}{2026}, pp. \bibinfo{pages}{23--32}.
  \DOIprefix\doi{10.1007/978-3-032-06329-8_3}.
\bibitem[{Postema(2011)}]{postema11diagnostic}
\bibinfo{author}{M.~Postema}, \bibinfo{title}{Fundamentals of Medical
  Ultrasonics}, \bibinfo{edition}{1} ed., \bibinfo{publisher}{CRC Press},
  \bibinfo{address}{Florence}, \bibinfo{year}{2011}.
\bibitem[{Villar et~al.(2014)Villar, Papageorghiou, Pang, Ohuma, Ismail,
  Barros, Lambert, Carvalho, Jaffer, Bertino, Gravett, Altman, Purwar,
  Frederick, Noble, Victora, Bhutta, and Kennedy}]{villar14likeness}
\bibinfo{author}{J.~Villar}, \bibinfo{author}{A.~T. Papageorghiou},
  \bibinfo{author}{R.~Pang}, \bibinfo{author}{E.~O. Ohuma},
  \bibinfo{author}{L.~C. Ismail}, \bibinfo{author}{F.~C. Barros},
  \bibinfo{author}{A.~Lambert}, \bibinfo{author}{M.~Carvalho},
  \bibinfo{author}{Y.~A. Jaffer}, \bibinfo{author}{E.~Bertino},
  \bibinfo{author}{M.~G. Gravett}, \bibinfo{author}{D.~G. Altman},
  \bibinfo{author}{M.~Purwar}, \bibinfo{author}{I.~O. Frederick},
  \bibinfo{author}{J.~A. Noble}, \bibinfo{author}{C.~G. Victora},
  \bibinfo{author}{Z.~A. Bhutta}, \bibinfo{author}{S.~H. Kennedy},
\newblock \bibinfo{title}{The likeness of fetal growth and newborn size across
  non-isolated populations in the intergrowth-21st project: the fetal growth
  longitudinal study and newborn cross-sectional study},
\newblock \bibinfo{journal}{The Lancet Diabetes \& Endocrinology}
  \bibinfo{volume}{2} (\bibinfo{year}{2014}) \bibinfo{pages}{781--792}.
  \DOIprefix\doi{10.1016/S2213-8587(14)70121-4}.
\bibitem[{Orlando and Vitale(2022)}]{orlando2022ussimandsegm}
\bibinfo{author}{J.~I. Orlando}, \bibinfo{author}{S.~Vitale},
  \bibinfo{title}{{US Simulation \& Segmentation: Real and Synthetic Abdominal
  Ultrasound Scans with Manual Segmentations}}, \bibinfo{year}{2022}.
  \URLprefix \url{https://www.kaggle.com/datasets/ignaciorlando/ussimandsegm},
  \bibinfo{note}{kaggle dataset. Accessed: 2026-07-17}.
\bibitem[{Meng et~al.(2018)Meng, Baumgartner, Sinclair, Housden, Rajchl, Gomez,
  Hou, Toussaint, Zimmer, Tan, Matthew, Rueckert, Schnabel, and
  Kainz}]{meng18automatic}
\bibinfo{author}{Q.~Meng}, \bibinfo{author}{C.~Baumgartner},
  \bibinfo{author}{M.~Sinclair}, \bibinfo{author}{J.~Housden},
  \bibinfo{author}{M.~Rajchl}, \bibinfo{author}{A.~Gomez},
  \bibinfo{author}{B.~Hou}, \bibinfo{author}{N.~Toussaint},
  \bibinfo{author}{V.~Zimmer}, \bibinfo{author}{J.~Tan},
  \bibinfo{author}{J.~Matthew}, \bibinfo{author}{D.~Rueckert},
  \bibinfo{author}{J.~Schnabel}, \bibinfo{author}{B.~Kainz},
\newblock \bibinfo{title}{Automatic shadow detection in 2d ultrasound images},
\newblock in: \bibinfo{booktitle}{Data Driven Treatment Response Assessment and
  Preterm, Perinatal, and Paediatric Image Analysis},
  \bibinfo{publisher}{Springer International Publishing},
  \bibinfo{address}{Cham}, \bibinfo{year}{2018}, pp. \bibinfo{pages}{66--75}.
\bibitem[{Wang et~al.(2004)Wang, Bovik, Sheikh, and Simoncelli}]{wang04ssim}
\bibinfo{author}{Z.~Wang}, \bibinfo{author}{A.~Bovik},
  \bibinfo{author}{H.~Sheikh}, \bibinfo{author}{E.~Simoncelli},
\newblock \bibinfo{title}{Image quality assessment: from error visibility to
  structural similarity},
\newblock \bibinfo{journal}{IEEE Transactions on Image Processing}
  \bibinfo{volume}{13} (\bibinfo{year}{2004}) \bibinfo{pages}{600--612}.
  \DOIprefix\doi{10.1109/TIP.2003.819861}.
\bibitem[{Paszke et~al.(2019)Paszke, Gross, Massa, Lerer, Bradbury, Chanan,
  Killeen, Lin, Gimelshein, Antiga, Desmaison, Kopf, Yang, DeVito, Raison,
  Tejani, Chilamkurthy, Steiner, Fang, Bai, and Chintala}]{paszke19pytorch}
\bibinfo{author}{A.~Paszke}, \bibinfo{author}{S.~Gross},
  \bibinfo{author}{F.~Massa}, \bibinfo{author}{A.~Lerer},
  \bibinfo{author}{J.~Bradbury}, \bibinfo{author}{G.~Chanan},
  \bibinfo{author}{T.~Killeen}, \bibinfo{author}{Z.~Lin},
  \bibinfo{author}{N.~Gimelshein}, \bibinfo{author}{L.~Antiga},
  \bibinfo{author}{A.~Desmaison}, \bibinfo{author}{A.~Kopf},
  \bibinfo{author}{E.~Yang}, \bibinfo{author}{Z.~DeVito},
  \bibinfo{author}{M.~Raison}, \bibinfo{author}{A.~Tejani},
  \bibinfo{author}{S.~Chilamkurthy}, \bibinfo{author}{B.~Steiner},
  \bibinfo{author}{L.~Fang}, \bibinfo{author}{J.~Bai},
  \bibinfo{author}{S.~Chintala},
\newblock \bibinfo{title}{{PyTorch: An Imperative Style, High-Performance Deep
  Learning Library}},
\newblock in: \bibinfo{editor}{H.~Wallach}, \bibinfo{editor}{H.~Larochelle},
  \bibinfo{editor}{A.~Beygelzimer}, \bibinfo{editor}{F.~d'Alché Buc},
  \bibinfo{editor}{E.~Fox}, \bibinfo{editor}{R.~Garnett} (Eds.),
  \bibinfo{booktitle}{Advances in Neural Information Processing Systems 32},
  \bibinfo{publisher}{Curran Associates, Inc.}, \bibinfo{year}{2019}, pp.
  \bibinfo{pages}{8024--8035}.
\bibitem[{Kingma and Ba(2017)}]{kingma17adam}
\bibinfo{author}{D.~P. Kingma}, \bibinfo{author}{J.~Ba}, \bibinfo{title}{Adam:
  A method for stochastic optimization}, \bibinfo{year}{2017}. \URLprefix
  \url{https://arxiv.org/abs/1412.6980}.
  \href{http://arxiv.org/abs/1412.6980}{{\tt arXiv:1412.6980}}.
\bibitem[{Wyburd et~al.(2021)Wyburd, Hesse, Aliasi, Jenkinson, Papageorghiou,
  Haak, and Namburete}]{wyburd21assessment}
\bibinfo{author}{M.~K. Wyburd}, \bibinfo{author}{L.~S. Hesse},
  \bibinfo{author}{M.~Aliasi}, \bibinfo{author}{M.~Jenkinson},
  \bibinfo{author}{A.~T. Papageorghiou}, \bibinfo{author}{M.~C. Haak},
  \bibinfo{author}{A.~I.~L. Namburete},
\newblock \bibinfo{title}{Assessment of regional cortical development through
  fissure based gestational age estimation in 3d fetal ultrasound},
\newblock in: \bibinfo{editor}{C.~H. Sudre}, \bibinfo{editor}{R.~Licandro},
  \bibinfo{editor}{C.~Baumgartner}, \bibinfo{editor}{A.~Melbourne},
  \bibinfo{editor}{A.~Dalca}, \bibinfo{editor}{J.~Hutter},
  \bibinfo{editor}{R.~Tanno}, \bibinfo{editor}{E.~Abaci~Turk},
  \bibinfo{editor}{K.~Van~Leemput}, \bibinfo{editor}{J.~Torrents~Barrena},
  \bibinfo{editor}{W.~M. Wells}, \bibinfo{editor}{C.~Macgowan} (Eds.),
  \bibinfo{booktitle}{Uncertainty for Safe Utilization of Machine Learning in
  Medical Imaging, and Perinatal Imaging, Placental and Preterm Image
  Analysis}, \bibinfo{publisher}{Springer International Publishing},
  \bibinfo{address}{Cham}, \bibinfo{year}{2021}, pp. \bibinfo{pages}{242--252}.
\bibitem[{Hesse et~al.(2024)Hesse, Dinsdale, and Namburete}]{hesse24prototype}
\bibinfo{author}{L.~S. Hesse}, \bibinfo{author}{N.~K. Dinsdale},
  \bibinfo{author}{A.~I.~L. Namburete},
\newblock \bibinfo{title}{Prototype learning for explainable brain age
  prediction},
\newblock in: \bibinfo{booktitle}{Proceedings of the IEEE/CVF Winter Conference
  on Applications of Computer Vision (WACV)}, \bibinfo{year}{2024}, pp.
  \bibinfo{pages}{7903--7913}.
\bibitem[{He et~al.(2016)He, Zhang, Ren, and Sun}]{he16deep}
\bibinfo{author}{K.~He}, \bibinfo{author}{X.~Zhang}, \bibinfo{author}{S.~Ren},
  \bibinfo{author}{J.~Sun},
\newblock \bibinfo{title}{Deep residual learning for image recognition},
\newblock in: \bibinfo{booktitle}{Proceedings of the IEEE Conference on
  Computer Vision and Pattern Recognition (CVPR)}, \bibinfo{year}{2016}.
\bibitem[{Meng et~al.(2019)Meng, Sinclair, Zimmer, Hou, Rajchl, Toussaint,
  Oktay, Schlemper, Gomez, Housden, Matthew, Rueckert, Schnabel, and
  Kainz}]{meng2019weakly}
\bibinfo{author}{Q.~Meng}, \bibinfo{author}{M.~Sinclair},
  \bibinfo{author}{V.~Zimmer}, \bibinfo{author}{B.~Hou},
  \bibinfo{author}{M.~Rajchl}, \bibinfo{author}{N.~Toussaint},
  \bibinfo{author}{O.~Oktay}, \bibinfo{author}{J.~Schlemper},
  \bibinfo{author}{A.~Gomez}, \bibinfo{author}{J.~Housden},
  \bibinfo{author}{J.~Matthew}, \bibinfo{author}{D.~Rueckert},
  \bibinfo{author}{J.~A. Schnabel}, \bibinfo{author}{B.~Kainz},
\newblock \bibinfo{title}{Weakly supervised estimation of shadow confidence
  maps in fetal ultrasound imaging},
\newblock \bibinfo{journal}{IEEE Transactions on Medical Imaging}
  \bibinfo{volume}{38} (\bibinfo{year}{2019}) \bibinfo{pages}{2755--2767}.
  \DOIprefix\doi{10.1109/TMI.2019.2913311}.
\bibitem[{Yesilkaynak et~al.(2024)Yesilkaynak, Duque, Wysocki, Velikova,
  Mateus, and Navab}]{yesilkaynak2024ultrasound}
\bibinfo{author}{V.~B. Yesilkaynak}, \bibinfo{author}{V.~G. Duque},
  \bibinfo{author}{M.~Wysocki}, \bibinfo{author}{Y.~Velikova},
  \bibinfo{author}{D.~Mateus}, \bibinfo{author}{N.~Navab},
  \bibinfo{title}{Ultrasound Confidence Maps with Neural Implicit
  Representation}, \bibinfo{publisher}{Springer Nature Switzerland},
  \bibinfo{year}{2024}, pp. \bibinfo{pages}{89--100}. \URLprefix
  \url{http://dx.doi.org/10.1007/978-3-031-66958-3\_7}.
  \DOIprefix\doi{10.1007/978-3-031-66958-3\_7}.
\bibitem[{Lundberg and Lee(2017)}]{lundberg17unified}
\bibinfo{author}{S.~M. Lundberg}, \bibinfo{author}{S.-I. Lee},
\newblock \bibinfo{title}{A unified approach to interpreting model
  predictions},
\newblock in: \bibinfo{booktitle}{Proceedings of the 31st International
  Conference on Neural Information Processing Systems}, NIPS'17,
  \bibinfo{publisher}{Curran Associates Inc.}, \bibinfo{address}{Red Hook, NY,
  USA}, \bibinfo{year}{2017}, p. \bibinfo{pages}{4768–4777}.

\end{thebibliography}

\newpage
\appendix
\section{Additional Results}
\label{apdx:A:additional-results}

\subsection{Additional Numerical Results}
\label{apdx:A:additional_numerical_results}

\begin{table}[!htbp]
\caption{Mean intensity difference $\Delta\mu$ across datasets for the original ($O$), Hughes shadow-reduced ($H$), and \proposed ($SR$) volumes when all measures are computed with the \proposed shadow maps. Values are shown as mean $\pm$ standard deviation. The two right-most columns report the subject-wise differences relative to \proposed. Differences marked with $^*$ are significantly different from 0 in a t-test with a p-value $<\frac{0.01}{6}=0.001667$ for multiple-comparison correction.}
\label{tab:appendix:numres_rf}
\centering
\resizebox{\linewidth}{!}{%
\begin{tabular}{@{}l|ccc|cc@{}}
\toprule
Dataset & $\Delta\mu_O$ & $\Delta\mu_H$ & $\Delta\mu_{SR}$ & $\left(\Delta\mu_O - \Delta\mu_{SR}\right)$ & $\left(\Delta\mu_H - \Delta\mu_{SR}\right)$ \\
\midrule
Syn liver & $64.12 \pm 27.21$ & $13.44 \pm 24.46$ & $16.93 \pm 17.47$ & $47.19 \pm 20.06^*$ & $-3.49 \pm 8.75^*$ \\
Abdomen & $87.65 \pm 16.84$ & $27.91 \pm 11.24$ & $8.04 \pm 6.41$ & $79.61 \pm 18.41^*$ & $19.87 \pm 11.25^*$ \\
Fetal Brain & $129.04 \pm 7.59$ & $58.90 \pm 18.16$ & $41.94 \pm 11.73$ & $87.10 \pm 12.57^*$ & $16.95 \pm 19.33^*$ \\
\bottomrule
\end{tabular}%
}
\end{table}

\begin{table}[tbp]
\caption{Mean intensity difference $\Delta\mu$ across datasets for the original ($O$), Hughes shadow-reduced ($H$), and \proposed ($SR$) volumes when all measures are computed with the Hughes shadow maps. Values are shown as mean $\pm$ standard deviation. The two right-most columns report the subject-wise differences relative to \proposed. Differences marked with $^*$ are significantly different from 0 in a t-test with a p-value $<\frac{0.01}{6}=0.001667$ for multiple-comparison correction.}
\label{tab:appendix:numres_hu}
\centering
\resizebox{\linewidth}{!}{%
\begin{tabular}{@{}l|ccc|cc@{}}
\toprule
Dataset & $\Delta\mu_O$ & $\Delta\mu_H$ & $\Delta\mu_{SR}$ & $\left(\Delta\mu_O - \Delta\mu_{SR}\right)$ & $\left(\Delta\mu_H - \Delta\mu_{SR}\right)$ \\
\midrule
Syn liver & $62.23 \pm 9.37$ & $13.61 \pm 7.37$ & $22.21 \pm 5.80$ & $40.02 \pm 6.35^*$ & $-8.59 \pm 8.81^*$ \\
Abdomen & $70.26 \pm 22.04$ & $13.41 \pm 10.13$ & $11.48 \pm 7.90$ & $58.78 \pm 21.19^*$ & $1.93 \pm 10.99$ \\
Fetal Brain & $88.01 \pm 12.29$ & $48.26 \pm 22.84$ & $15.25 \pm 11.92$ & $72.75 \pm 15.33^*$ & $33.00 \pm 24.52^*$ \\
\bottomrule
\end{tabular}%
}
\end{table}

\FloatBarrier
\subsection{Additional Results Age Prediction}
\label{apdx:A:additional_age_pred}

\begin{figure*}[!htbp]
\centering
\edef\figwidth{0.32} 
\begin{subfigure}{\figwidth\textwidth}
\centering
  \includegraphics[width=\linewidth]{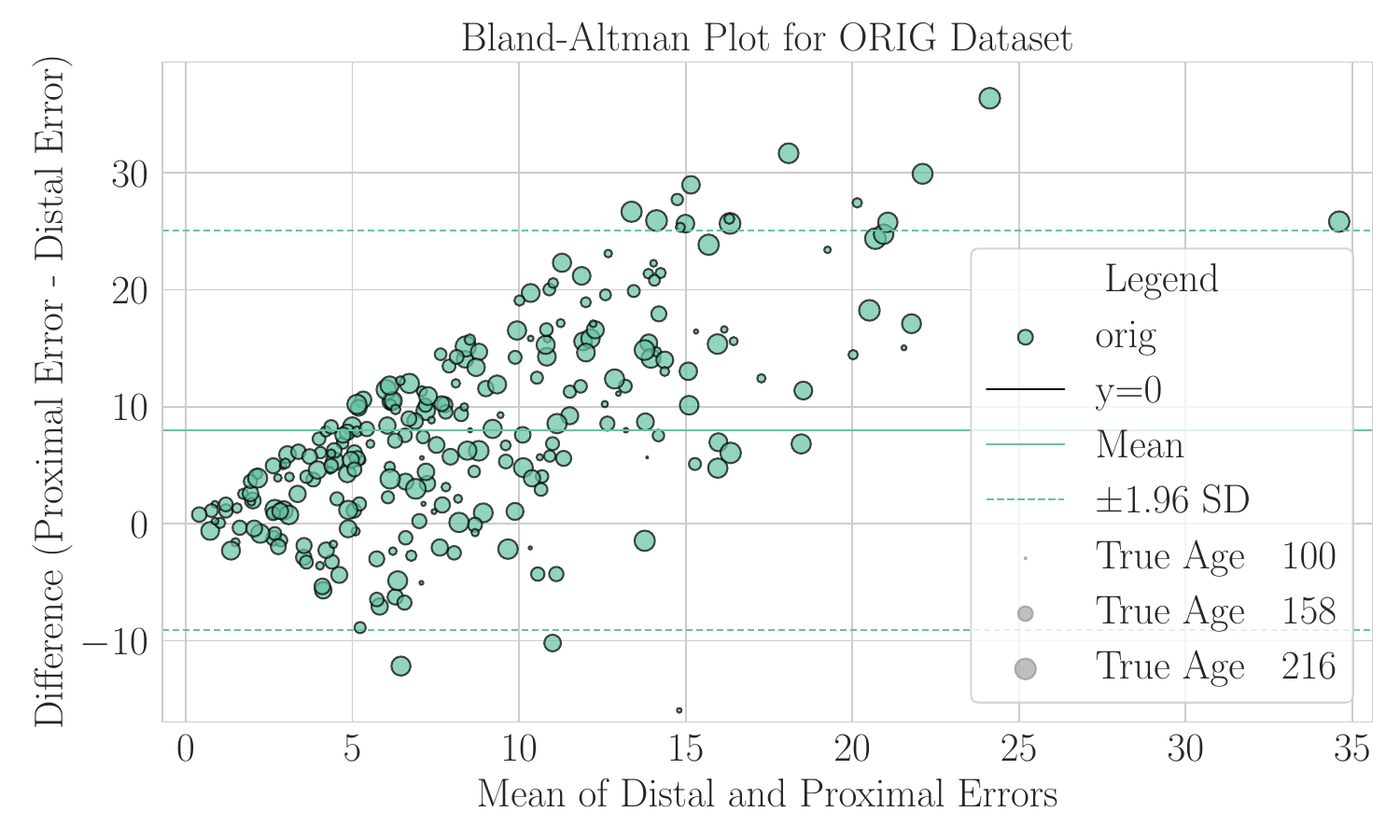}
  \caption{Bland--Altman plot for the original images}
  \label{fig:appendix:agepred:orig}
\end{subfigure}
\hfill
\begin{subfigure}{\figwidth\textwidth}
\centering
  \includegraphics[width=\linewidth]{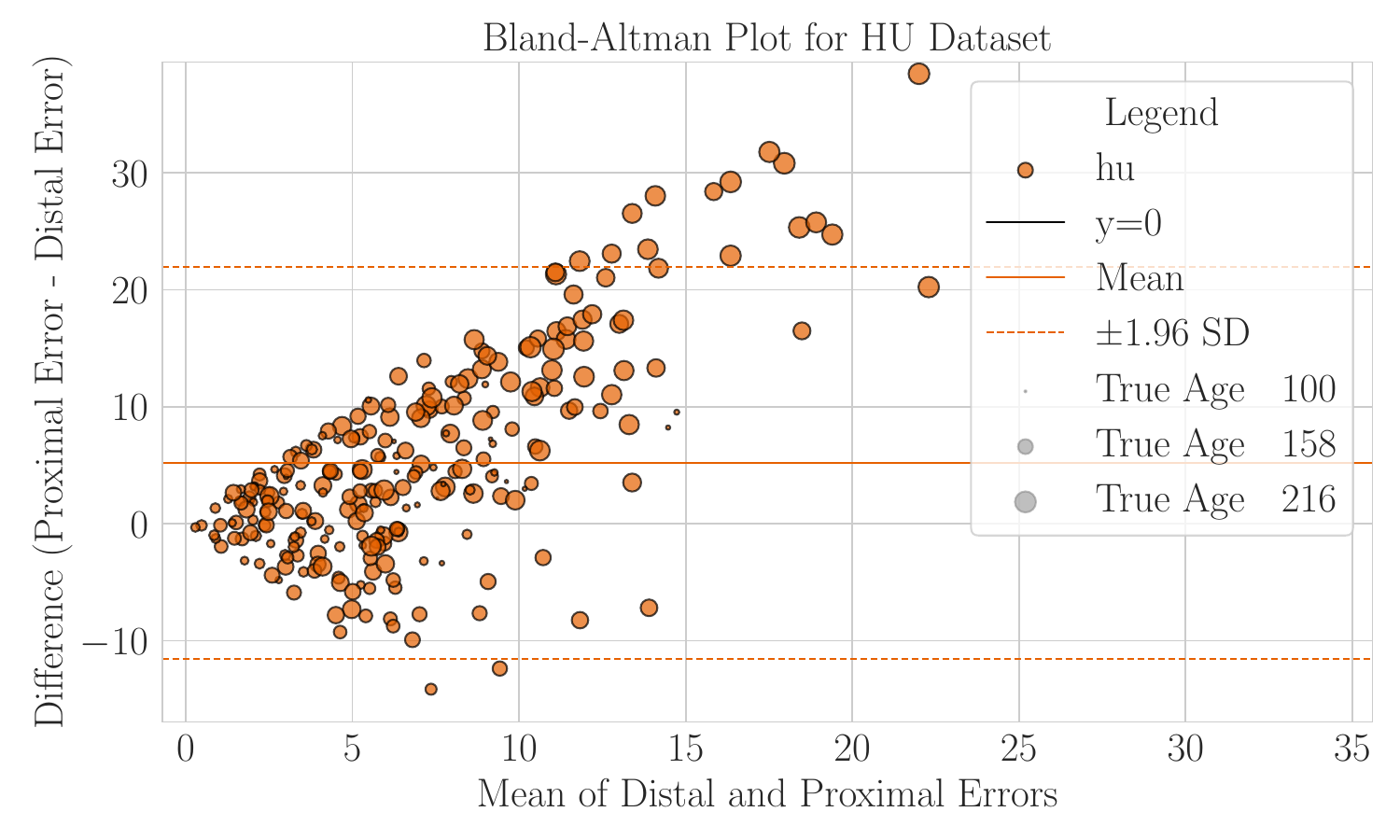}
  \caption{Bland--Altman plot for the Hughes-reduced images}
  \label{fig:appendix:agepred:hu}
\end{subfigure}
\hfill
\begin{subfigure}{\figwidth\textwidth}
\centering
  \includegraphics[width=\linewidth]{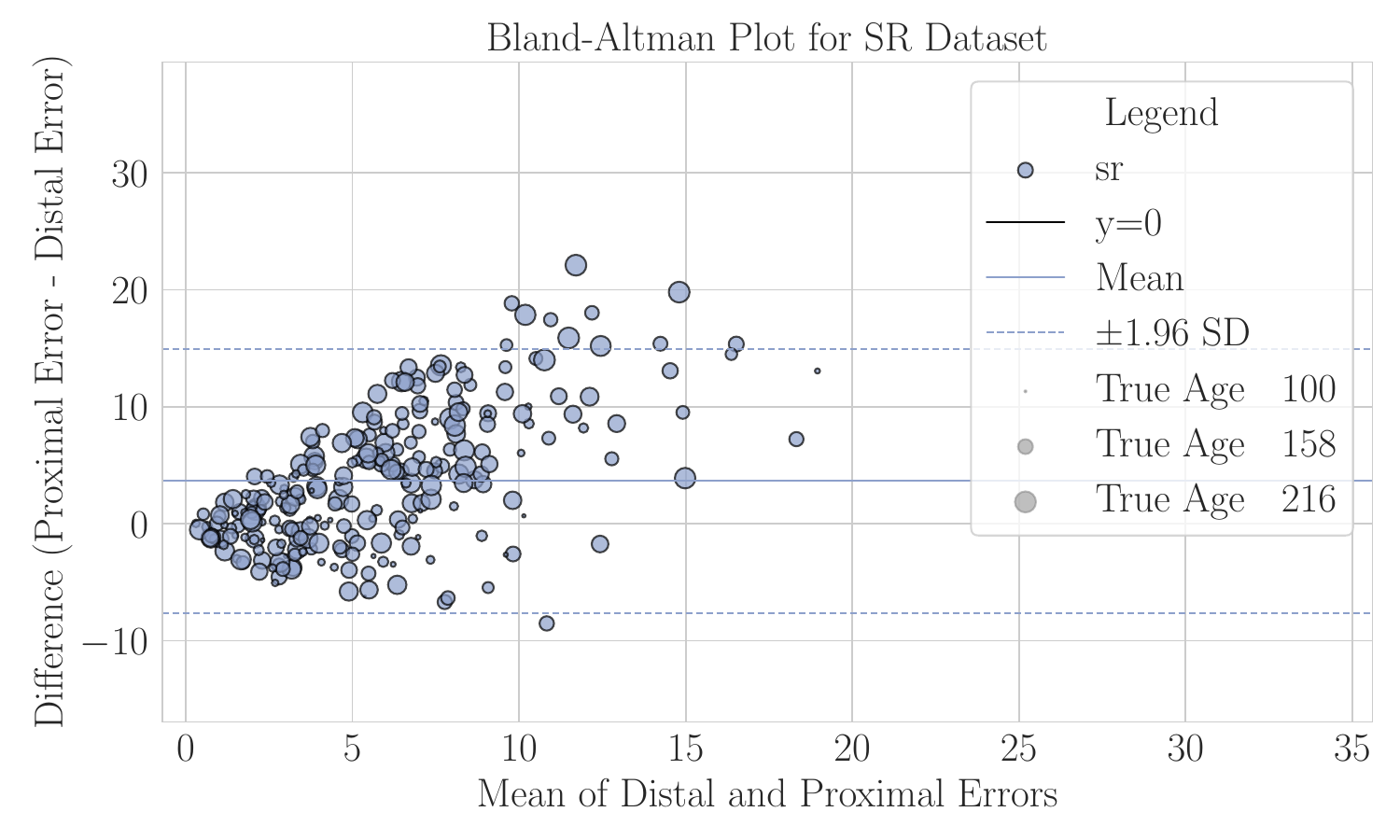}
  \caption{Bland--Altman plot for the \proposed images}
  \label{fig:appendix:agepred:sr}
\end{subfigure}
\caption{This figure complements \Cref{fig:results:bland_altman} by showing the individual data points underlying the convex hulls in that figure.}
\label{fig:appendix:agepred}
\end{figure*}
\FloatBarrier

\section{Ablation Studies}
\label{apdx:B:ablation-studies}

\subsection{Sensitivity with respect to sampling in simulation space}
\label{apdx:B:sens:sampling}

\begin{figure*}[!htbp]
\centering
\edef\figwidth{0.32} 
\begin{subfigure}{\figwidth\textwidth}
\centering
  \includegraphics[width=\linewidth]{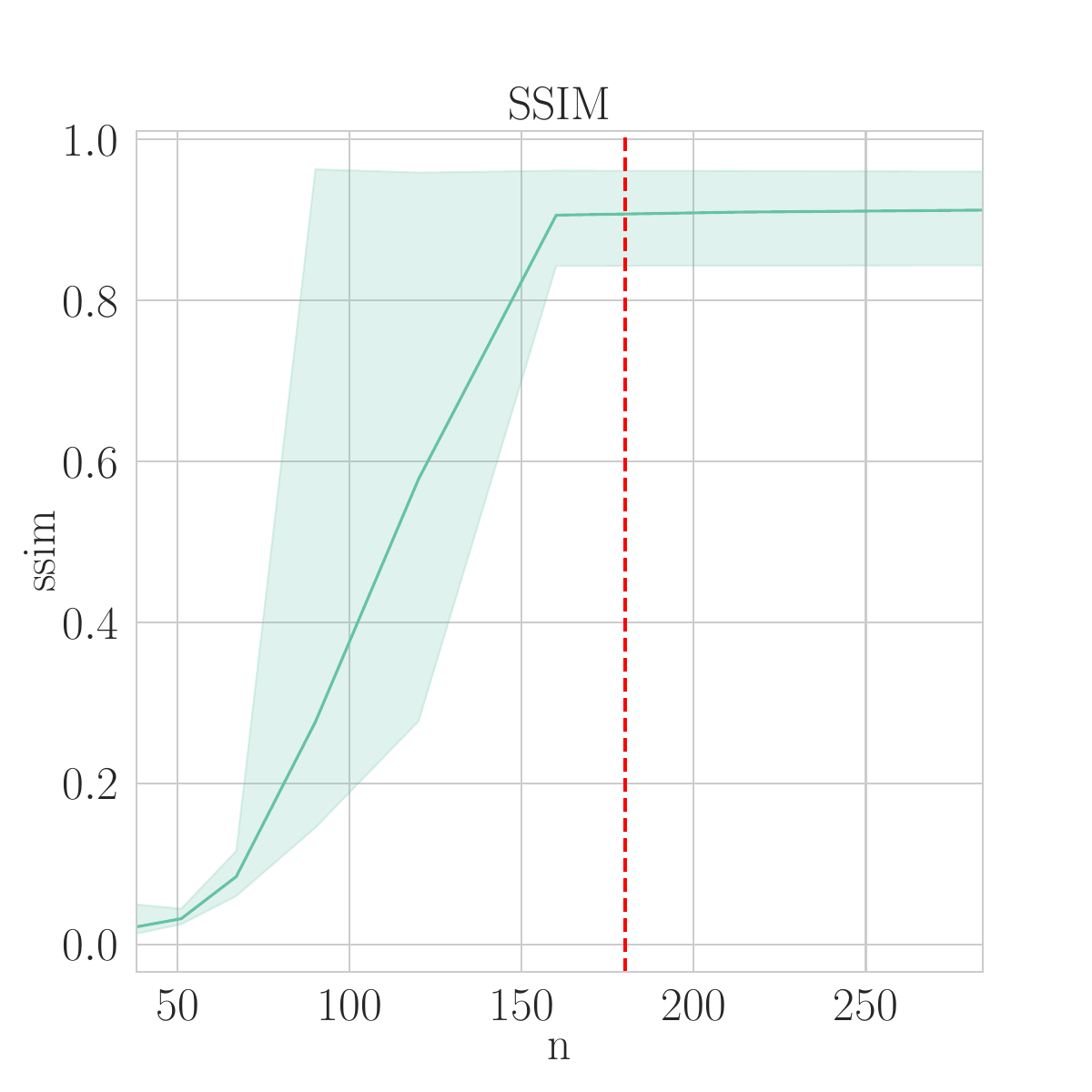}
  \caption{Scaling of SSIM over $n$.}
  \label{fig:apdx:B:sampling_num:ssim}
\end{subfigure}
\hfill
\begin{subfigure}{\figwidth\textwidth}
\centering
  \includegraphics[width=\linewidth]{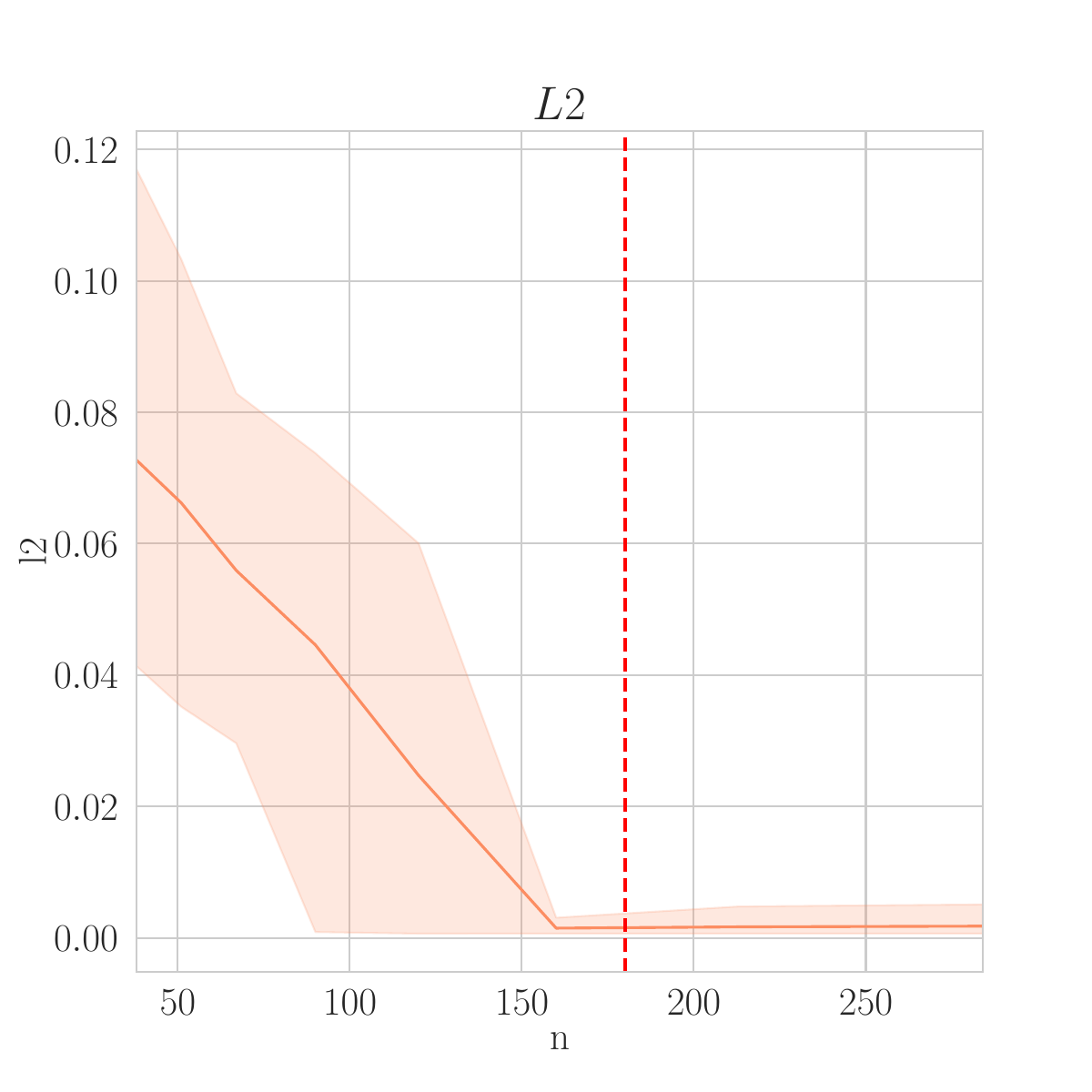}
  \caption{Scaling of $L_2$ over $n$.}
  \label{fig:apdx:B:sampling_num:l2}
\end{subfigure}
\hfill
\begin{subfigure}{\figwidth\textwidth}
\centering
  \includegraphics[width=\linewidth]{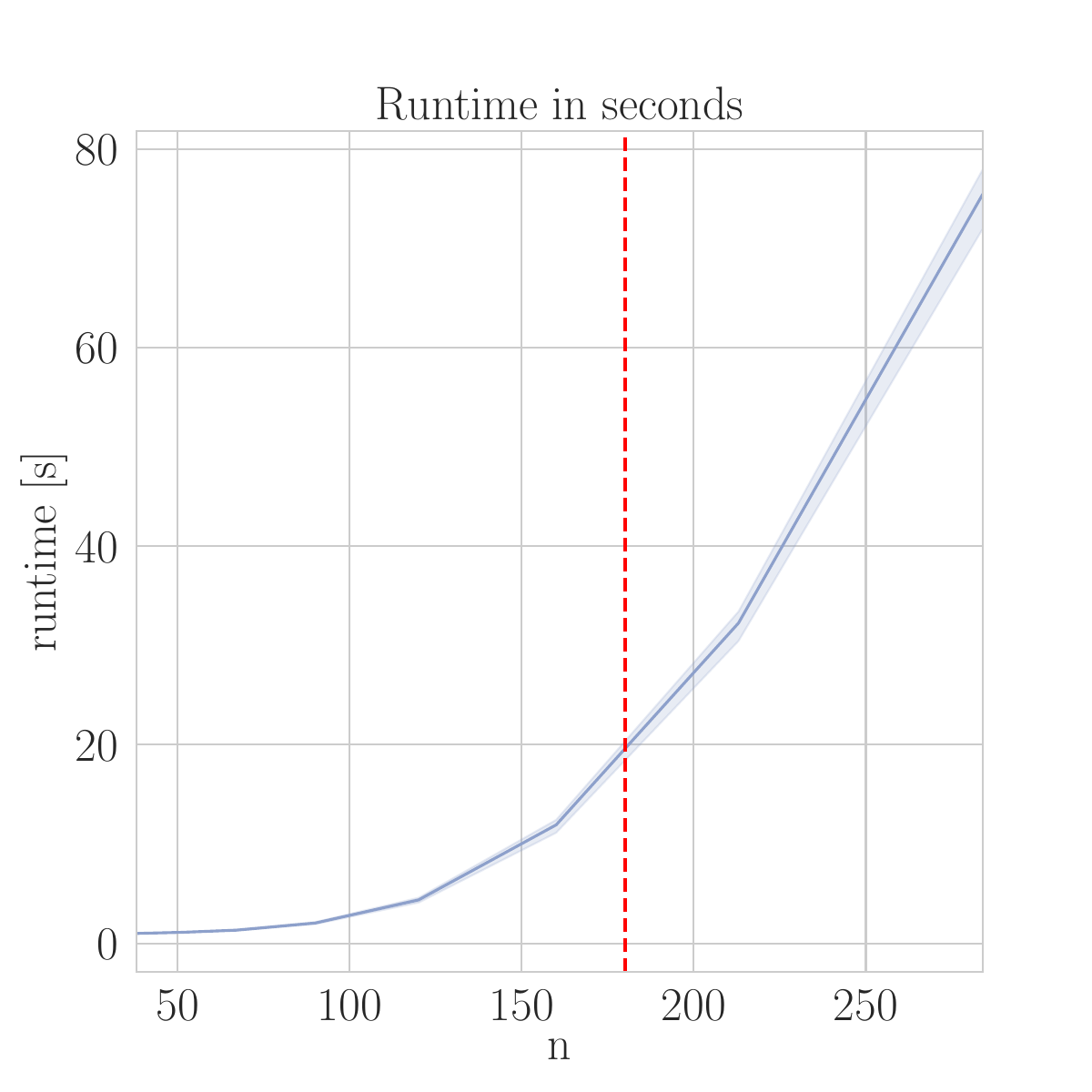}
  \caption{Scaling of runtime over $n$.}
  \label{fig:apdx:B:sampling_num:rt}
\end{subfigure}
\caption{Similarity measures between simulated and original volumes as a function of $n$, where $n$ is the edge length of the sampled cube in simulation space. The red line indicates the chosen value.}
\label{fig:appendix:sampling_num}
\end{figure*}

\begin{figure*}[!htbp]
\centering
\edef\figwidth{0.24} 
\begin{subfigure}{\figwidth\textwidth}
\centering
  \scalebox{1}[-1]{\includegraphics[width=\linewidth]{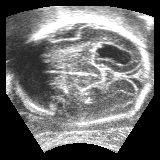}}
  \caption{orig}
  \label{fig:apdx:B:sampling_vis:orig}
\end{subfigure}
\hfill
\begin{subfigure}{\figwidth\textwidth}
\centering
  \scalebox{1}[-1]{\includegraphics[width=\linewidth]{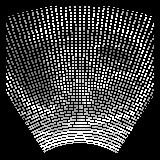}}
  \caption{n=51}
  \label{fig:apdx:B:sampling_vis:51}
\end{subfigure}
\hfill
\begin{subfigure}{\figwidth\textwidth}
\centering
  \scalebox{1}[-1]{\includegraphics[width=\linewidth]{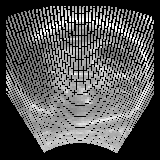}}
  \caption{n=67}
  \label{fig:apdx:B:sampling_vis:67}
\end{subfigure}
\hfill
\begin{subfigure}{\figwidth\textwidth}
\centering
  \scalebox{1}[-1]{\includegraphics[width=\linewidth]{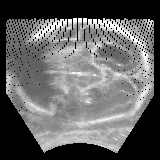}}
  \caption{n=90}
  \label{fig:apdx:B:sampling_vis:90}
\end{subfigure}
\hfill
\begin{subfigure}{\figwidth\textwidth}
\centering
  \scalebox{1}[-1]{\includegraphics[width=\linewidth]{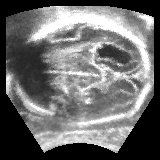}}
  \caption{n=120}
  \label{fig:apdx:B:sampling_vis:120}
\end{subfigure}
\hfill
\begin{subfigure}{\figwidth\textwidth}
\centering
  \scalebox{1}[-1]{\includegraphics[width=\linewidth]{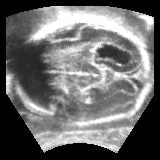}}
  \caption{n=160}
  \label{fig:apdx:B:sampling_vis:160}
\end{subfigure}
\hfill
\begin{subfigure}{\figwidth\textwidth}
\centering
  \scalebox{1}[-1]{\includegraphics[width=\linewidth]{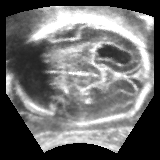}}
  \caption{n=213}
  \label{fig:apdx:B:sampling_vis:213}
\end{subfigure}
\hfill
\begin{subfigure}{\figwidth\textwidth}
\centering
  \scalebox{1}[-1]{\includegraphics[width=\linewidth]{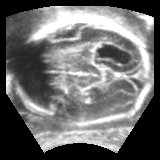}}
  \caption{n=284}
  \label{fig:apdx:B:sampling_vis:284}
\end{subfigure}
\caption{This figure shows example volumes reconstructed using different densities of sampling in simulation space.}
\label{fig:apdx:B:sampling_vis}
\end{figure*}

\FloatBarrier
\subsection{Sensitivity to initial values}
\label{apdx:B:sens:init}

\begin{figure*}[!htbp]
\centering
\edef\figwidth{0.48} 
\begin{subfigure}{\figwidth\textwidth}
\centering
  \includegraphics[width=\linewidth]{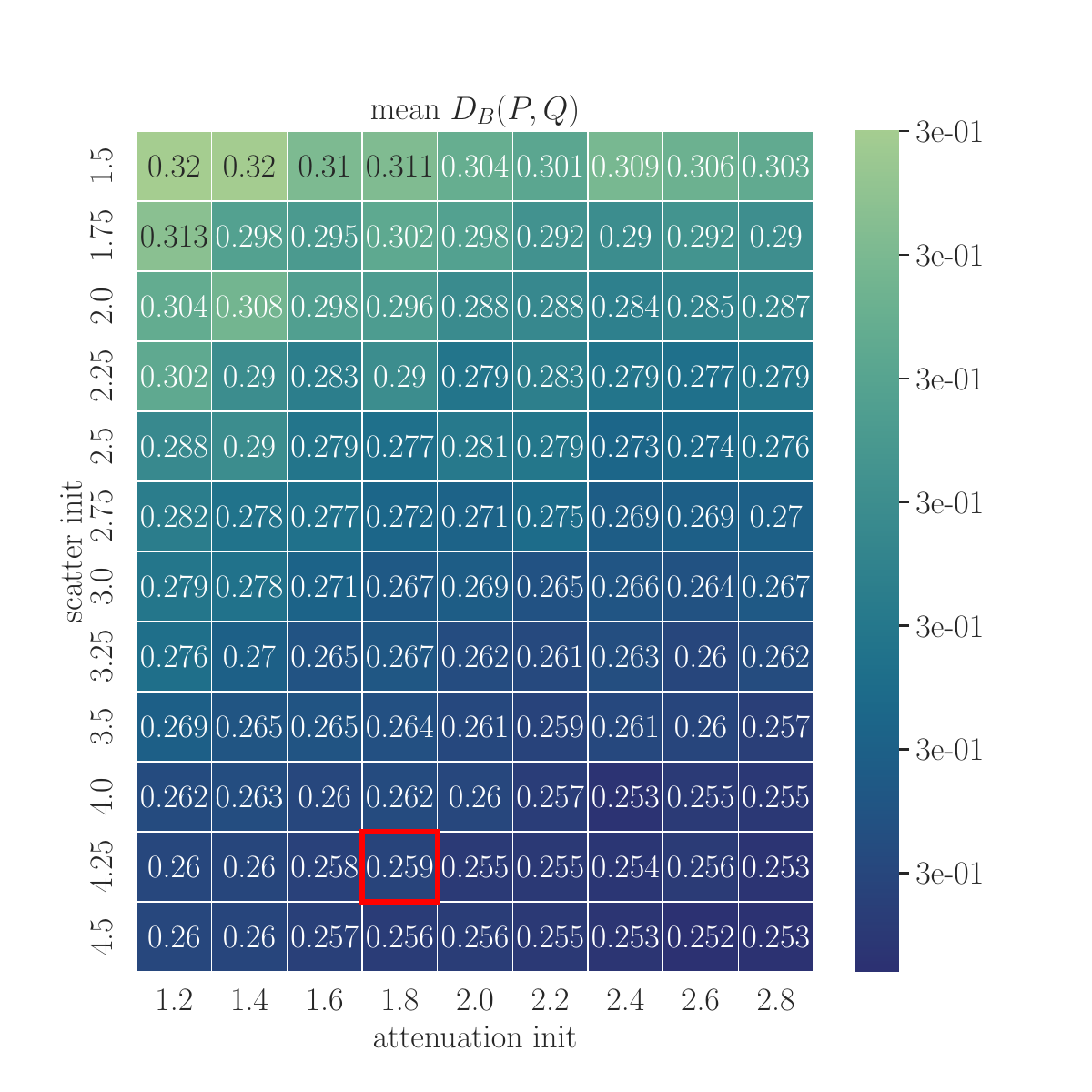}
  \caption{Heat map $D_{B}(P,Q)$ over the tested initial values}
  \label{fig:apdx:B:init:bhat}
\end{subfigure}
\hfill
\begin{subfigure}{\figwidth\textwidth}
\centering
  \includegraphics[width=\linewidth]{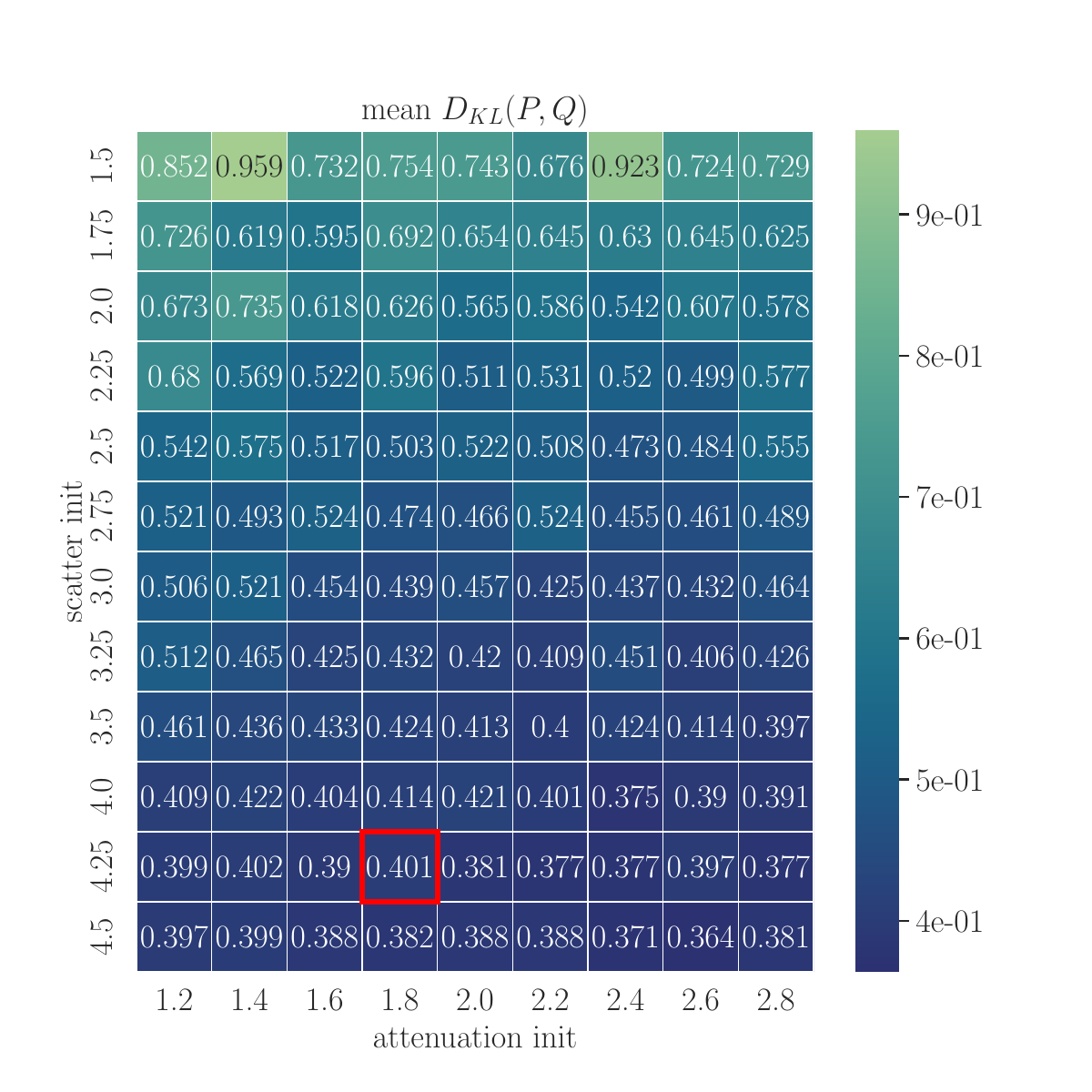}
  \caption{Heat map $D_{KL}(P,Q)$ over the tested initial values}
  \label{fig:apdx:B:init:kl}
\end{subfigure}
\hfill
\begin{subfigure}{\figwidth\textwidth}
\centering
  \includegraphics[width=\linewidth]{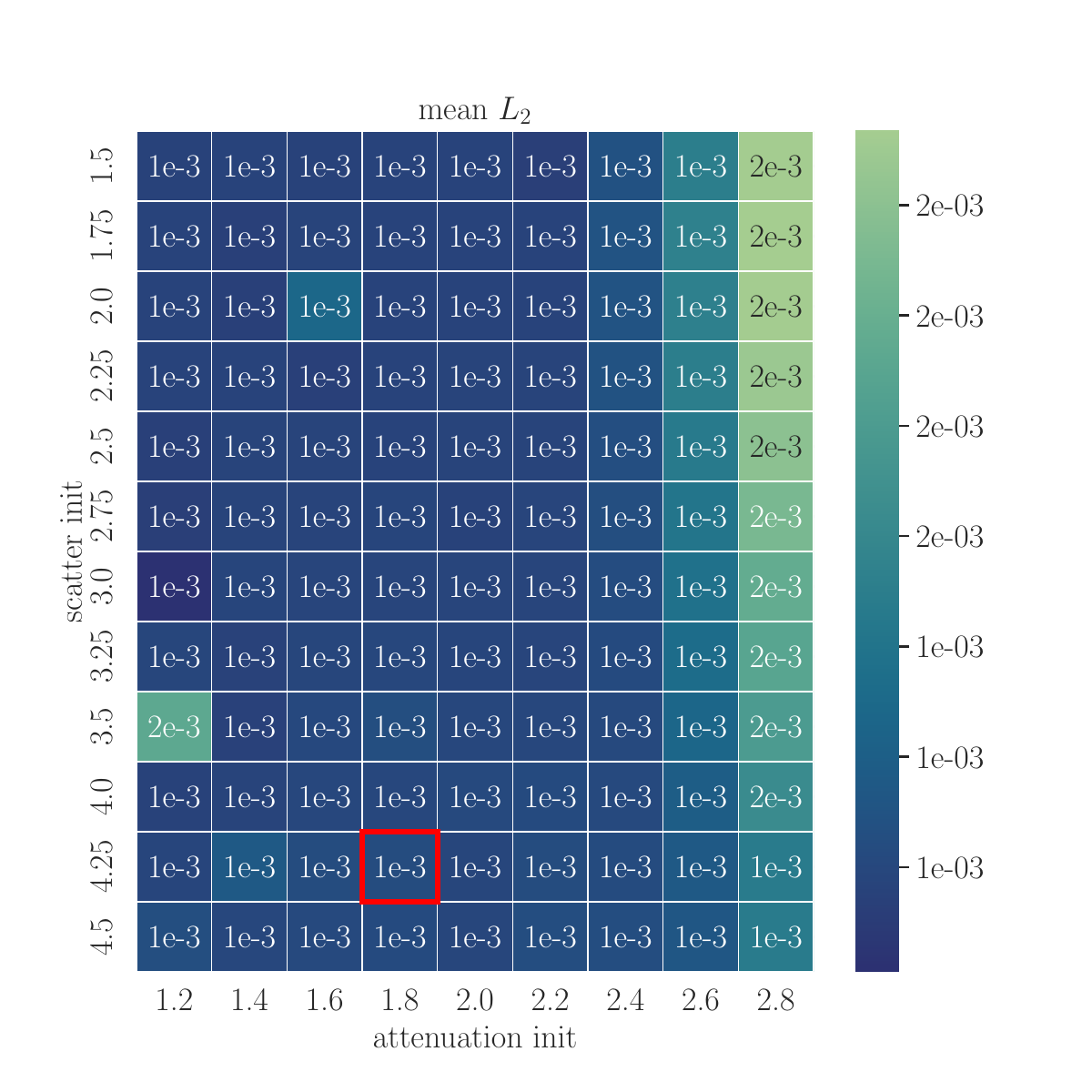}
  \caption{Heat map $L_2$ over the tested initial values}
  \label{fig:apdx:B:init:l2}
\end{subfigure}
\hfill
\begin{subfigure}{\figwidth\textwidth}
\centering
  \includegraphics[width=\linewidth]{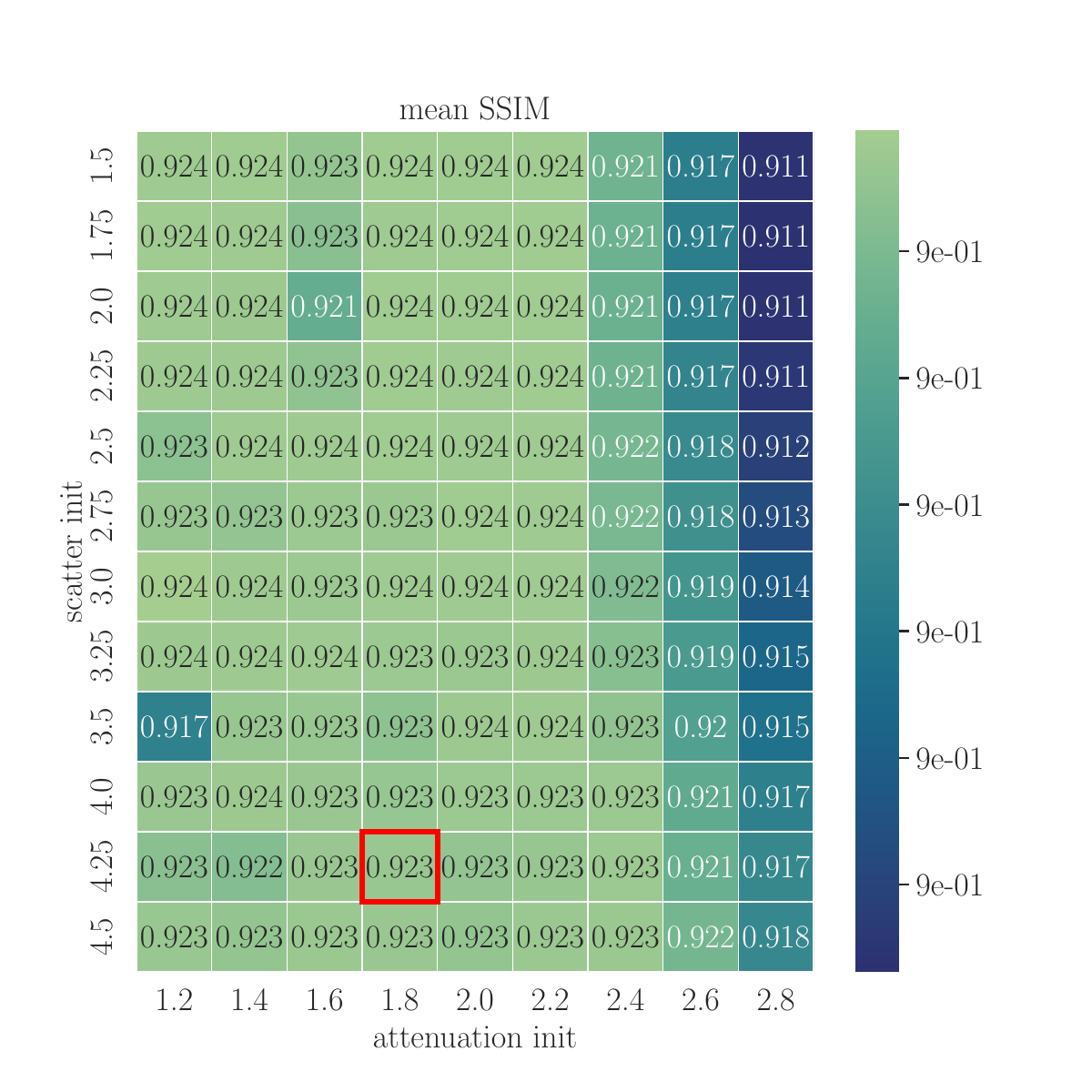}
  \caption{Heat map SSIM over the tested initial values}
  \label{fig:apdx:B:init:ssim}
\end{subfigure}
\caption{Heat maps of shadow-reduction measures for different combinations of initial values. The red box marks the selected combination.}
\label{fig:apdx:B:init}
\end{figure*}

\FloatBarrier
\subsection{Sensitivity with respect to loss function weighting}
\label{apdx:B:sens:loss}

\begin{figure}[!htbp]
\centering
\includegraphics[width=.5\linewidth]{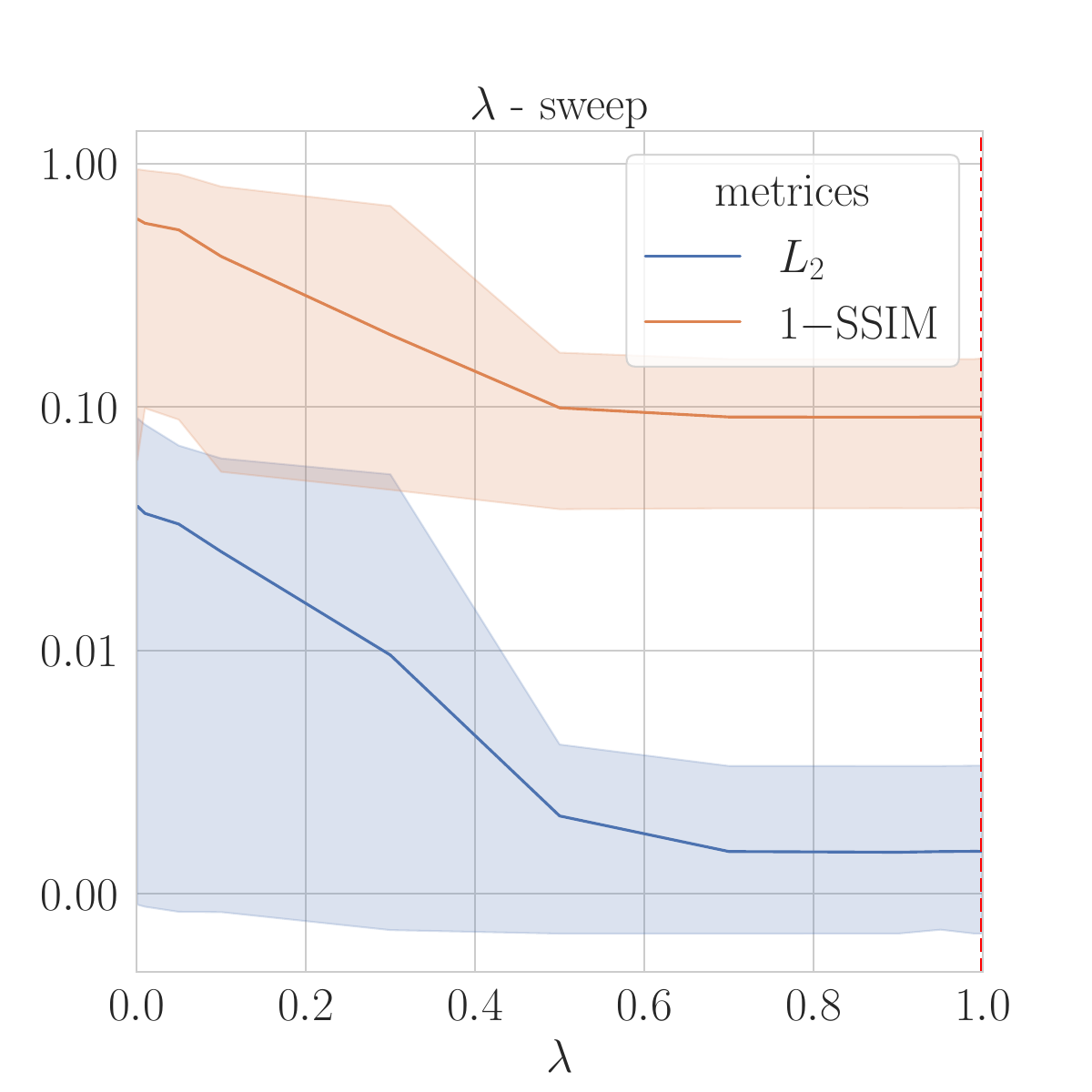}
\caption{SSIM and $L_2$ after convergence for different weighting factors. The red line indicates the chosen value.}
\label{fig:apdx:B:lambda}
\end{figure}

\FloatBarrier
\subsection{Convergence behaviour}
\label{apdx:B:sens:conv}

\begin{figure}[!htbp]
\centering
\includegraphics[width=.5\linewidth]{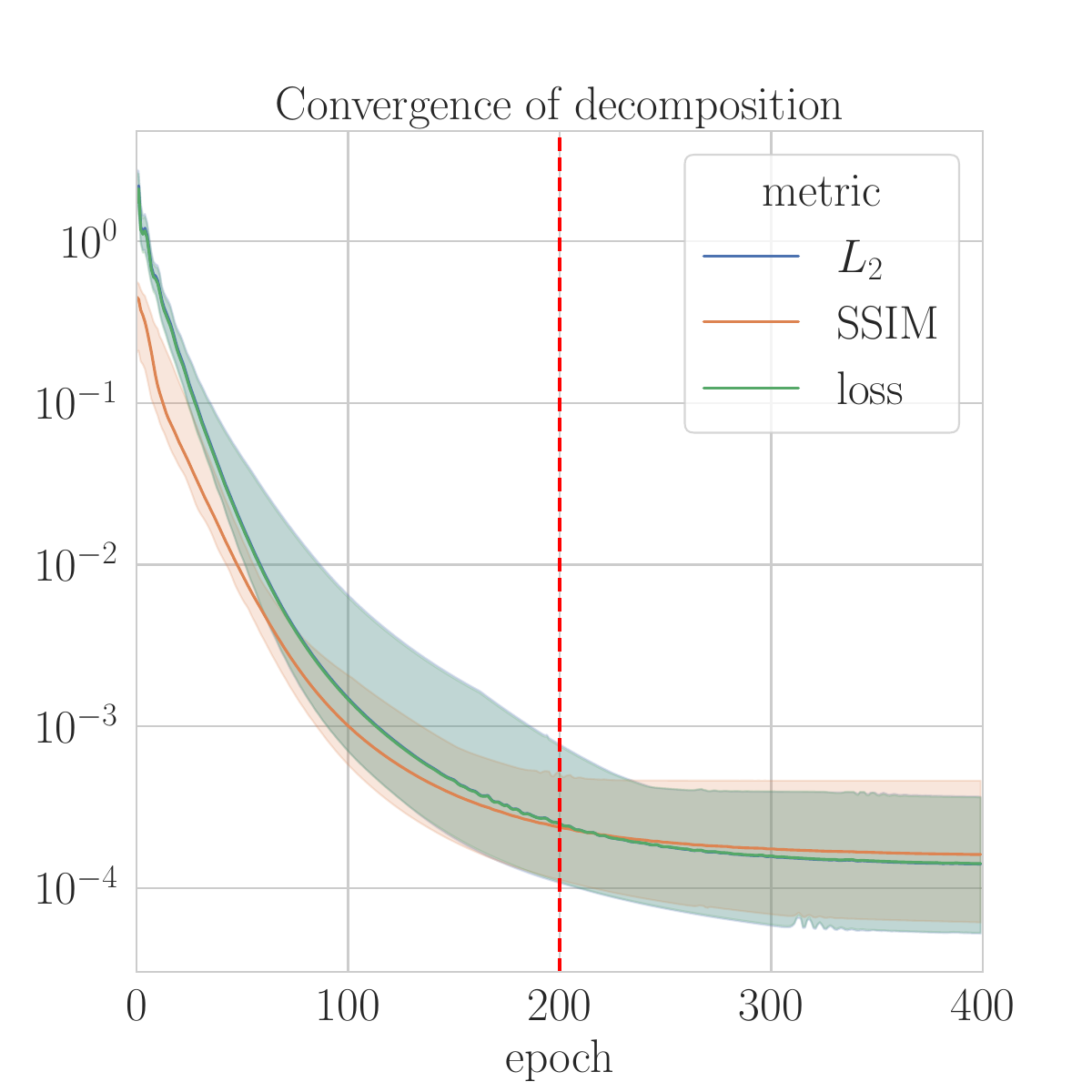}
\caption{Convergence of the $L_2$ and SSIM metrics over epochs. The red line indicates the chosen value.}
\label{fig:apdx:B:conv}
\end{figure}

\FloatBarrier
\subsection{Sensitivity of the shadow metric to the number of histogram buckets}
\label{apdx:B:sens:buckets}

\begin{figure*}[!htbp]
\centering
\edef\figwidth{0.32} 
\begin{subfigure}{\figwidth\textwidth}
\centering
  \includegraphics[width=\linewidth]{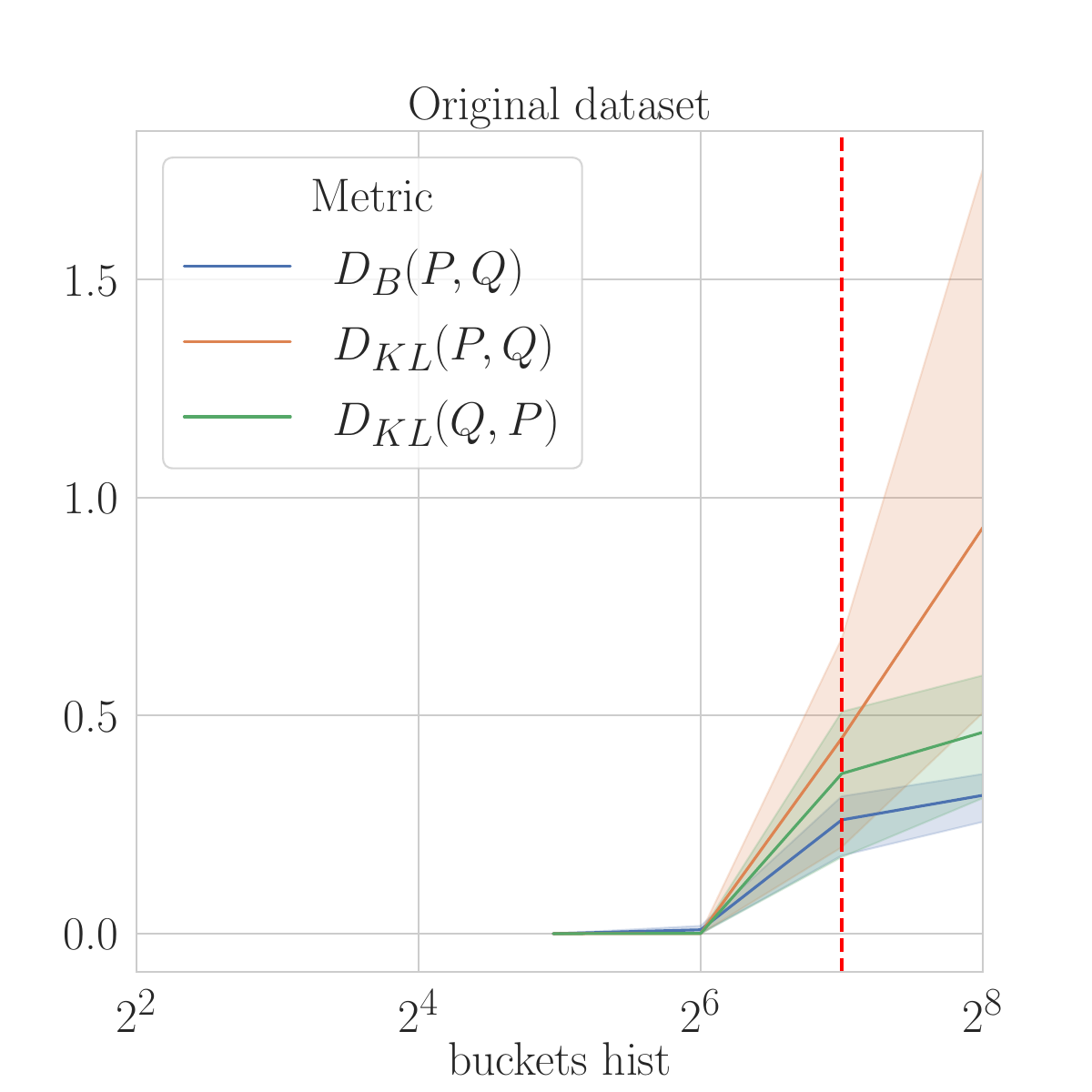}
\caption{Scaling of the three metrics as a function of the number of buckets on the original dataset.}
  \label{fig:apdx:B:buckets:orig}
\end{subfigure}
\hfill
\begin{subfigure}{\figwidth\textwidth}
\centering
  \includegraphics[width=\linewidth]{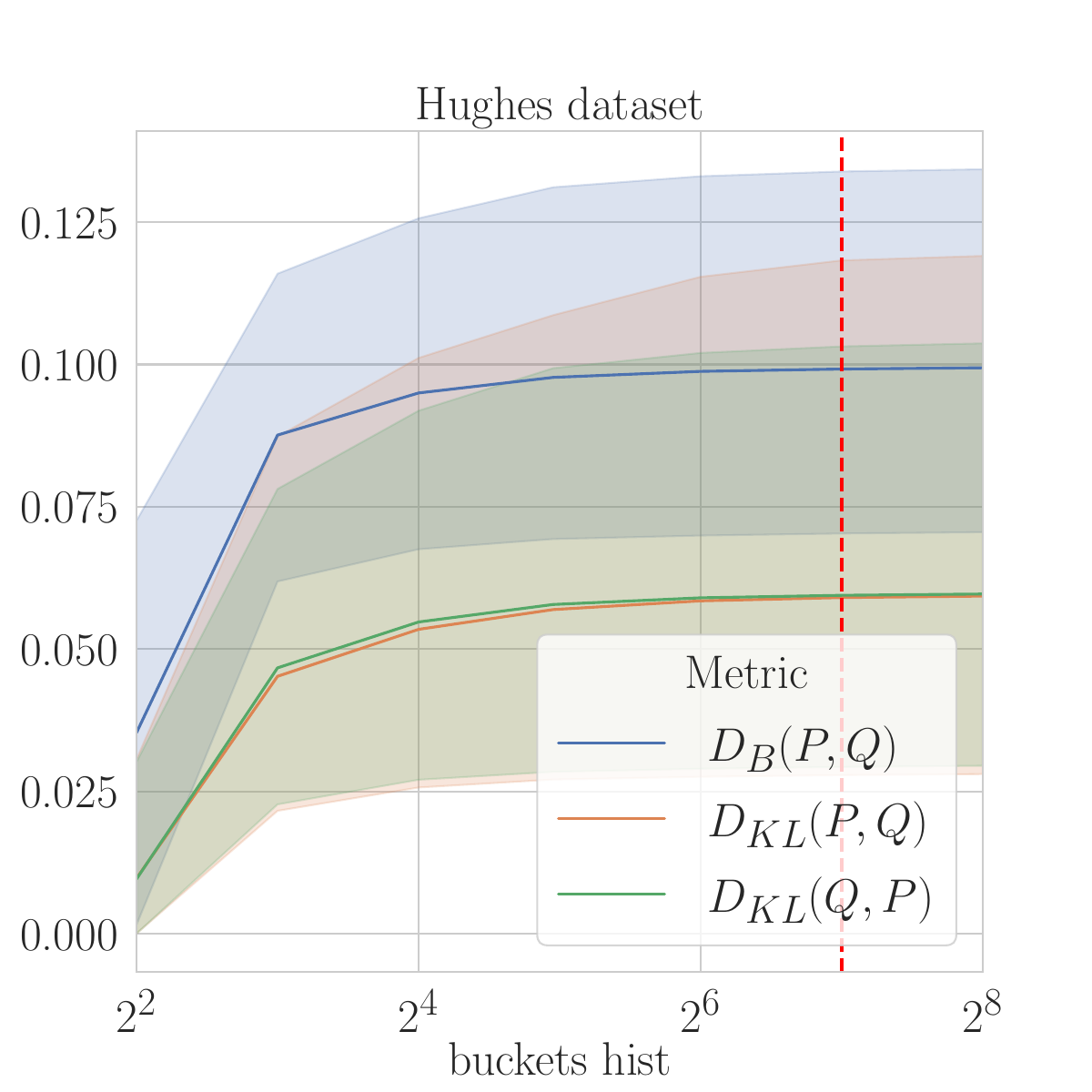}
\caption{Scaling of the three metrics as a function of the number of buckets on the Hughes dataset.}
  \label{fig:apdx:B:buckets:hu}
\end{subfigure}
\hfill
\begin{subfigure}{\figwidth\textwidth}
\centering
  \includegraphics[width=\linewidth]{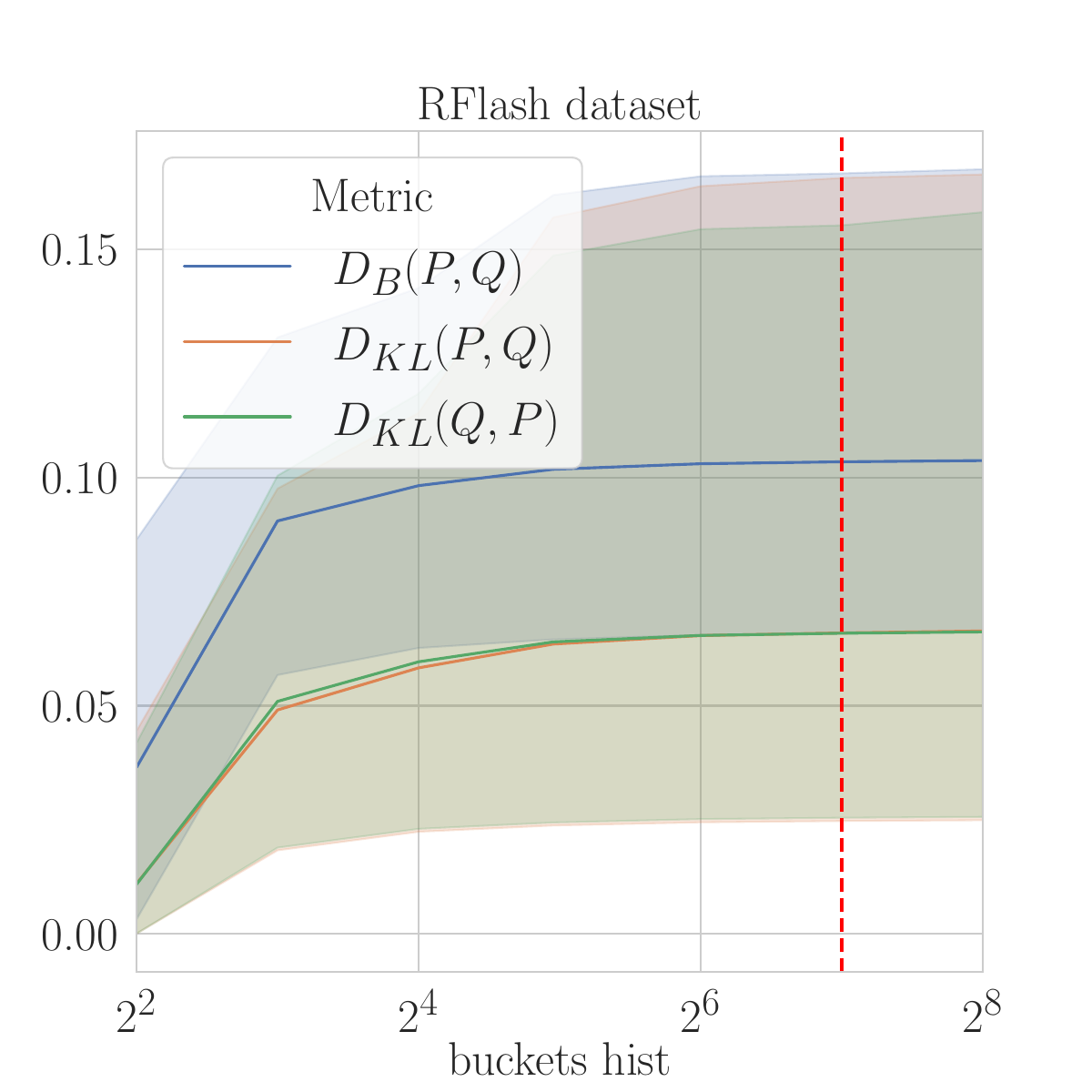}
\caption{Scaling of the three metrics as a function of the number of buckets on the \proposed dataset.}
  \label{fig:apdx:B:buckets:rf}
\end{subfigure}
\caption{Mean and confidence intervals for the three metrics described above as a function of the number of buckets used to calculate the histograms. The red line indicates the chosen value.}
\label{fig:apdx:B:buckets}
\end{figure*}
\FloatBarrier

\end{document}